\documentclass[conference]{IEEEtran}
\usepackage{times}

\usepackage[numbers]{natbib}
\usepackage{multicol}
\usepackage[bookmarks=true]{hyperref}

\usepackage{amsmath,amsfonts}
\usepackage{array}
\usepackage[caption=false,font=normalsize,labelfont=sf,textfont=sf]{subfig}
\usepackage{textcomp}
\usepackage{stfloats}
\usepackage{url}
\usepackage{verbatim}
\usepackage{graphicx}

\usepackage[font=footnotesize,labelfont=bf]{caption}
\usepackage{hyperref}
\usepackage{float}
\makeatletter
\let\newfloat\newfloat@ltx
\makeatother

\usepackage[ruled,vlined]{algorithm2e}
\usepackage{amssymb}  
\usepackage{color}
\usepackage[dvipsnames]{xcolor}
\usepackage{mathtools}
\usepackage[utf8]{inputenc}
\usepackage{xr}
\usepackage{mathrsfs}
\usepackage{enumerate}
\usepackage{multicol}

\usepackage{changepage}
\usepackage{mdframed}
\usepackage{xcolor,soul}
\soulregister\cite7
\soulregister\ref7
\soulregister\eqref7
\soulregister\aligned7
\colorlet{usercolorname}{red!00} 
\sethlcolor{usercolorname}

\mdfdefinestyle{MyFrame}{%
    linecolor=black,
    outerlinewidth=2pt,
    innertopmargin=4pt,
    innerbottommargin=4pt,
    innerrightmargin=4pt,
    innerleftmargin=4pt,
        leftmargin = 4pt,
        rightmargin = 4pt
        }
        
\mdfdefinestyle{MyFrame_Skip}{%
    linecolor=black,
    outerlinewidth=2pt,
    skipabove=12pt,
    skipbelow=12pt,
    innertopmargin=4pt,
    innerbottommargin=4pt,
    innerrightmargin=4pt,
    innerleftmargin=4pt,
        leftmargin = 4pt,
        rightmargin = 4pt
        }

\DeclareMathOperator*{\argmin}{arg\,min}

\newtheorem{theorem}{Theorem}
\newtheorem{remark}{Remark}

\begin{document}

\title{Hamilton-Jacobi Reachability Analysis for Hybrid Systems with Controlled and Forced Transitions}




%
\author{\authorblockN{Javier Borquez,
Shuang Peng,
Yiyu Chen, 
Quan Nguyen and
Somil Bansal}
\authorblockA{ University of Southern California, LA, USA. 
E-mail: \{javierbo, shuangpe, yiyuc, quann, somilban\}@usc.edu.}
}

\makeatletter
\let\@oldmaketitle\@maketitle
    \renewcommand{\@maketitle}{\@oldmaketitle
    \centering
    \vspace{1em}
    \includegraphics[width=1.00\textwidth]{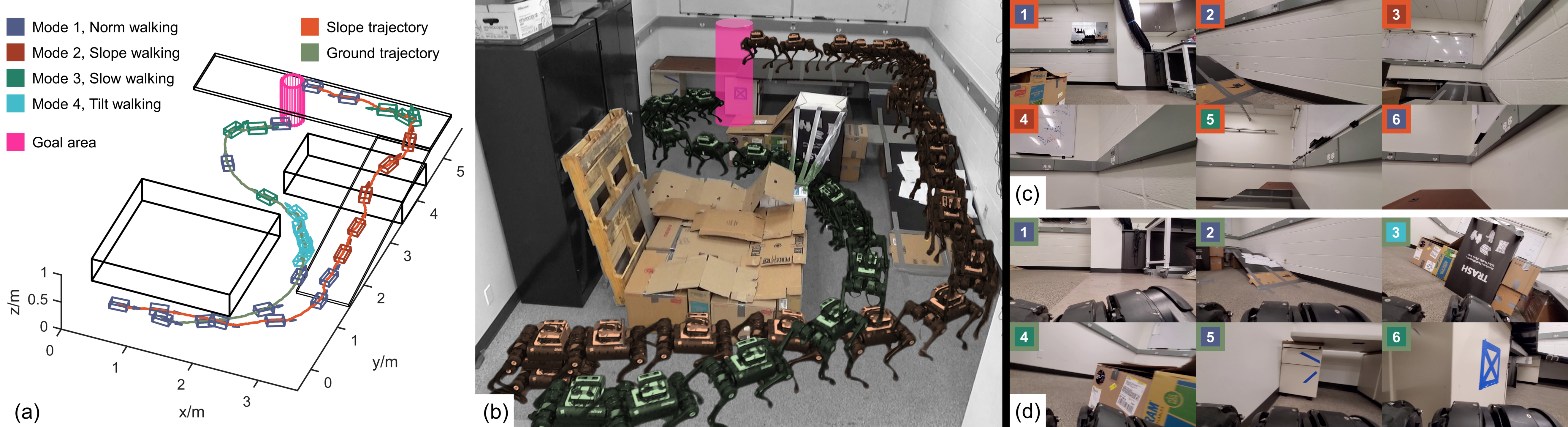}
    \captionof{figure}{We propose a generalized Hamilton-Jacobi (HJ) reachability framework to compute \hl{Backward Reachable Tubes (BRT)} and safe controllers for hybrid dynamical systems with nonlinear dynamics and discrete transitions.
    Here, we apply the proposed framework for optimal mode planning for a quadruped robot that switches between different walking gaits (modeled as discrete modes in the hybrid system) in order to reach its goal safely (i.e., without colliding with any obstacles on the way).
    We compute the BRT of the robot and obtain the optimal continuous control and mode sequence from the BRT. 
    We show the corresponding quadruped trajectories in (a). The quadruped safely navigates through the slope or narrow ground obstacles to reach the goal area (pink cylinder). The goal area is independent of the $z$ position, so the robot can reach it via slope or ground route.
    Different colors along the robot trajectory correspond to different walking modes that the robot autonomously chooses to avoid collision while navigating as quickly as possible to the goal. (b) Overlaid trajectories in the real-world obstacle setup. (c, d) First-person view along the slope route (c) and the ground route (d). Videos of the experiments and source materials can be found on the project website \href{https://javierborquez.github.io/HybridReachability/}{\textbf{javierborquez.github.io/HybridReachability}}.}
    \vspace{-0.2cm} 
    \label{fig:exp_main_result}
    \setcounter{figure}{1}
  }
\makeatother

\maketitle

\begin{abstract}

Hybrid dynamical systems with nonlinear dynamics are one of the most general modeling tools for representing robotic systems, especially contact-rich systems. 
However, providing guarantees regarding the safety or performance of nonlinear hybrid systems remains a challenging problem because it requires simultaneous reasoning about continuous state evolution and discrete mode switching.
In this work, we address this problem by extending classical Hamilton-Jacobi (HJ) reachability analysis, a formal verification method for continuous-time nonlinear dynamical systems, to hybrid dynamical systems. 
We characterize the reachable sets for hybrid systems through a generalized value function defined over discrete and continuous states of the hybrid system.
We also provide a numerical algorithm to compute this value function and obtain the reachable set.
Our framework can compute reachable sets for hybrid systems consisting of multiple discrete modes, each with its own set of nonlinear continuous dynamics, discrete transitions that can be directly commanded or forced by a discrete control input, while still accounting for control bounds and adversarial disturbances in the state evolution. 
Along with the reachable set, the proposed framework also provides an optimal continuous and discrete controller to ensure system safety.
We demonstrate our framework in several simulation case studies, as well as on a real-world testbed to solve the optimal mode planning problem for a quadruped with multiple gaits.
\end{abstract}

\IEEEpeerreviewmaketitle

\section{Introduction}
%
%
%
\vspace{-0.5em}
Hybrid dynamical systems are a popular and versatile tool to model robotic systems that exhibit both continuous and discrete dynamics \cite{goebel2009hybrid, johnson2016hybrid}.
They are ``hybrid'' in the sense that they contain both continuous and discrete state variables, wherein the continuous flow is interleaved with discrete events or jumps.
For example, for a legged robot, the swinging of a leg has continuous dynamics, whereas contacts with ground are well-modeled as discrete events.
However, as with many advanced modeling tools, their very complexity and richness presents challenges, particularly when it comes to ensuring safety and performance for such systems.
Designing safe controllers for hybrid dynamical systems demands simultaneous reasoning about the continuous evolution of states and the discrete mode transitions, a task that can quickly become computationally intensive and conceptually challenging \cite{kong2021salted}.

Hamilton-Jacobi (HJ) Reachability analysis is a powerful and effective approach for analyzing and controlling hybrid dynamical systems. 
It provides a comprehensive understanding of the system's behavior by characterizing the Backward Reachable Tube (BRT) of the system -- the set of all possible initial states that will eventually steer the system to some target states despite best control effort.
Thus, if the target set represents the undesirable states for the system, the BRT represents the set of states that are potentially unsafe for the system and should be avoided. 
The converse of the BRT thus provides a safe operation region for the system.
On the other hand, if the target set represents a desirable set of states, the BRT can similarly be computed to obtain the set of states from which the target set can eventually be reached under optimal control effort.
Along with the BRT, the reachability analysis also provides a safe controller for the system to keep the system in the safe region (respectively liveness controller if the target set is desirable).

The utility of HJ reachability lies in its ability to handle general nonlinear dynamics, control bounds, and dynamics uncertainty during the BRT computation \cite{bansal2017hamilton}.
Correspondingly, a number of methods have been developed to theoretically characterize and numerically compute BRTs for continuous-time dynamical systems (see \cite{bansal2017hamilton} for a survey).
However, similar frameworks are lacking for hybrid dynamical systems.
Existing reachability methods for hybrid dynamical systems either rely on heuristics to handle the effect of discrete transitions on the BRT or are limited to restrictive dynamics (e.g., linear dynamics) and transitions (e.g., linear guard sets). 

In this work, we propose a HJ reachability framework to compute BRTs for hybrid dynamical systems with general nonlinear dynamics and discrete transitions.
We pay special attention to two types of transitions between discrete modes that are particularly common in robotics applications: 1) \textit{Forced switches/transitions} - discrete state transitions that are forced because a particular set of continuous states were reached (also referred to as \textit{guard sets} in the literature). For example, a forced switch can be used to model the transition of a jumping robot from standing dynamics to ballistic flight dynamics when the leg stops being in contact with the ground.
2) \textit{Controlled switches/transitions} - discrete state transitions that can be activated (or not) at any moment while in a particular discrete operation mode, e.g., disengaging cruise control on a vehicle whenever the driver wants to take manual control of the speed. 
Underlying each discrete mode are continuous robot dynamics along with a continuous state and control input.

We propose an exact reachability operator for hybrid dynamical systems with nonlinear dynamics and controlled and forced transitions.
Our framework formulates the BRT computation for hybrid systems as a robust optimal control problem, wherein the continuous and discrete control inputs attempt to keep the system outside the target set.
Using the principle of dynamic programming, we \textit{simultaneously} reason about optimal continuous and discrete control inputs for this optimal control problem.
We show that the value function corresponding to this optimal control problem is given by a generalized version of Hamilton-Jacobi-Bellman (HJB) partial differential equation (PDE).
Intuitively, the value function describes how ``safe'' or ``unsafe'' a particular pair of discrete and continuous state is.
Once computed, the value function can be used to obtain the BRT for the hybrid system.
Furthermore, we demonstrate that the existing numerical algorithms to compute BRTs for continuous-time dynamical systems can easily be extended to compute this generalized value function, thereby providing both the BRT as well as the safe discrete and continuous control inputs for the hybrid system.
%
%

The proposed framework can handle hybrid systems consisting of multiple discrete modes with nonlinear dynamics, controlled switches, forced switches, nonlinear guard functions, (discontinuous) state resets upon discrete transitions, control bounds, as well as potential dynamics uncertainty.   
We demonstrate the efficacy of our framework in simulation and on a real-world quadruped testbed that transitions between different walking gaits (modeled as discrete modes) in order to safely reach its goal in a cluttered environment.
Our results illustrate that the proposed framework can simultaneously reason about continuous and discrete transitions to enhance system safety.
To summarize, the key contributions of this work are:
\begin{enumerate}
    \item A unified HJ reachability framework for hybrid systems that can handle nonlinear dynamics, guard functions, as well as forced and controlled discrete transitions;
    \item A numerical algorithm to compute the reachable tubes as well as safe controllers for hybrid dynamical systems;
    \item Demonstration of hybrid reachability framework for autonomous aircraft conflict resolution, a hopper robot, as well as safe navigation for a real-world quadruped testbed.
\end{enumerate}

\section{\label{related_work}Related Work}
Reachability analysis for hybrid dynamical systems has been extensively studied by both computer science and control communities. 
Correspondingly, a number of different approaches have been proposed in the literature. 
In this section, we provide a brief overview of some of the prominent approaches, as well as the current research gaps.

Historically, one of the most well-studied methods for hybrid reachabiilty analysis is rooted in hybrid automata theory, such as timed automata \cite{alur1994theory} and linear hybrid automata \cite{alur1991hybrid}.
Reachability computations in these methods are typically based on propagation of polygonal sets under constant rate dynamics. 
Tools for automatically performing these computations have also been developed \cite{henzinger1995user, maler1995synthesis, yovine1997kronos}.
However, these methods typically impose restrictive assumptions on the underlying continuous dynamics, such as limiting the analysis to linear dynamics and not allowing continuous control inputs, restricting their direct use in robotics applications. 

Other class of approaches extend reachability tools for continuous state and time dynamical systems to incorporate discrete switches \cite{deshpande1994viable, altin2020semicontinuity, chai2018forward, girard2013computational}. 
These approaches include zonotopes-based methods, computability theory, taylor models, Satisfiability Modulo Theory, Hamilton-Jacobi reachability, among others. 
For instance, zonotopes-based methods represent reachable sets as zonotopes or a mixture of zonotopes, and solve the hybrid reachability problem by propagating these sets and considering their interaction with discrete event transitions modeled as guard sets \cite{ zono_avoid_intersect_2012, zono_poly_2010, contact_zono_2023, nonlin_zono_2015}.
However, the efficient computation of reachable sets for
hybrid systems with nonlinear dynamics remains a difficult problem to solve.
Furthermore, it is challenging to account for controlled transitions in these methods, which is often a key requirement in the motion and trajectory planning of hybrid systems such as legged robots (e.g., where the system might want to transition between different discrete gaits in order to safely reach its goal).
%

Other reachability methods for hybrid systems rely on rigorous computable analysis theory to represent reachable sets as geometric objects \cite{ariadne_denotable_reach08,finite_time_denotable_2005}.
Taylor models have also been used in the analysis of hybrid reachable sets to provide rigorous enclosures of the set trajectories, while accounting for uncertainties and errors in the computation \cite{taylor_hyb_flow12,flow_star_2013,taylor_nonlin_guard_20}.
Satisfiability Modulo Theory (SMT) has also been used for the reachability analysis of hybrid systems.
Such methods encode dynamics and discrete mode transitions as first-order formulas over real numbers that are solved using an SMT solver \cite{large_discrete_2007, delta_reach2015}.
However, above methods typically compute an over-approximations of reachable sets, while limiting the discrete transitions to forced transitions.
Another approach for computing BRTs for hybrid dynamical systems is via Hamilton-Jacobi (HJ) Reachability analysis \cite{tomlin1996hybrid, lygeros1998controller, lygeros1996hierarchical}. Its advantages include compatibility with general non-linear system dynamics, formal treatment of bounded disturbances, and the ability to deal with state and input constraints \cite{bansal2017hamilton}. 
Several classical and modern works have addressed the control and safety analysis of hybrid dynamical systems through HJ Reachability \cite{coll_avoid_HJI_hybrid_2017,drone_backflip_2011,air_3modes_200l,air_7modes_1999}.  
\hl{These methods rely on an iterative algorithm to compute the BRT, wherein the BRT is iteratively refined in each discrete mode (using a continuous reach-avoid operator) based on the last computed BRT and the discrete predecessor maps. Intuitively, these predecessor maps capture the effect of forced and controlled transitions on the BRT of the system. This process is repeated until the BRT reaches a fixed point and converges. 
However, the BRT computation might require several iterations to converge and thus can be time-consuming.
Furthermore, the iterative reachability algorithm often requires hand-coding the discrete predecessor maps \cite{air_3modes_200l}, which can be challenging when the guard and reset maps are complex.}   
A recent work generalizes the HJ reachability framework to account for discontinuous state changes upon state resets~\cite{ROA_reset_2022}.
However, generalizing the HJ reachability framework for multiple discrete modes and accounting for controlled transitions still remains challenging.
We aim to overcome these limitations by proposing a generalization of traditional HJ reachability framework to account for discrete transitions (both forced and controlled) in the reachability computation. 
The proposed method provides an exact reachability operator for hybrid dynamical systems and can readily be applied to systems with nonlinear dynamics, control bounds, and dynamics uncertainty, without requiring any hand-coding.
\section{\label{problem}Problem Formulation}
\hl{We consider a hybrid dynamical system defined as:}
\begin{equation}\label{eq:hyb_def}
H=((Q \times X),(U \times D), f, \operatorname{Inv}, \Sigma, R),
\end{equation}
\hl{where $Q := \{q_1, q_2, \hdots, q_N\}$ is a finite set of discrete modes. We also refer to $q_i$ as the discrete state of the system.}
\begin{figure}[h!]
\begin{center} 
\includegraphics[width=0.95\columnwidth]{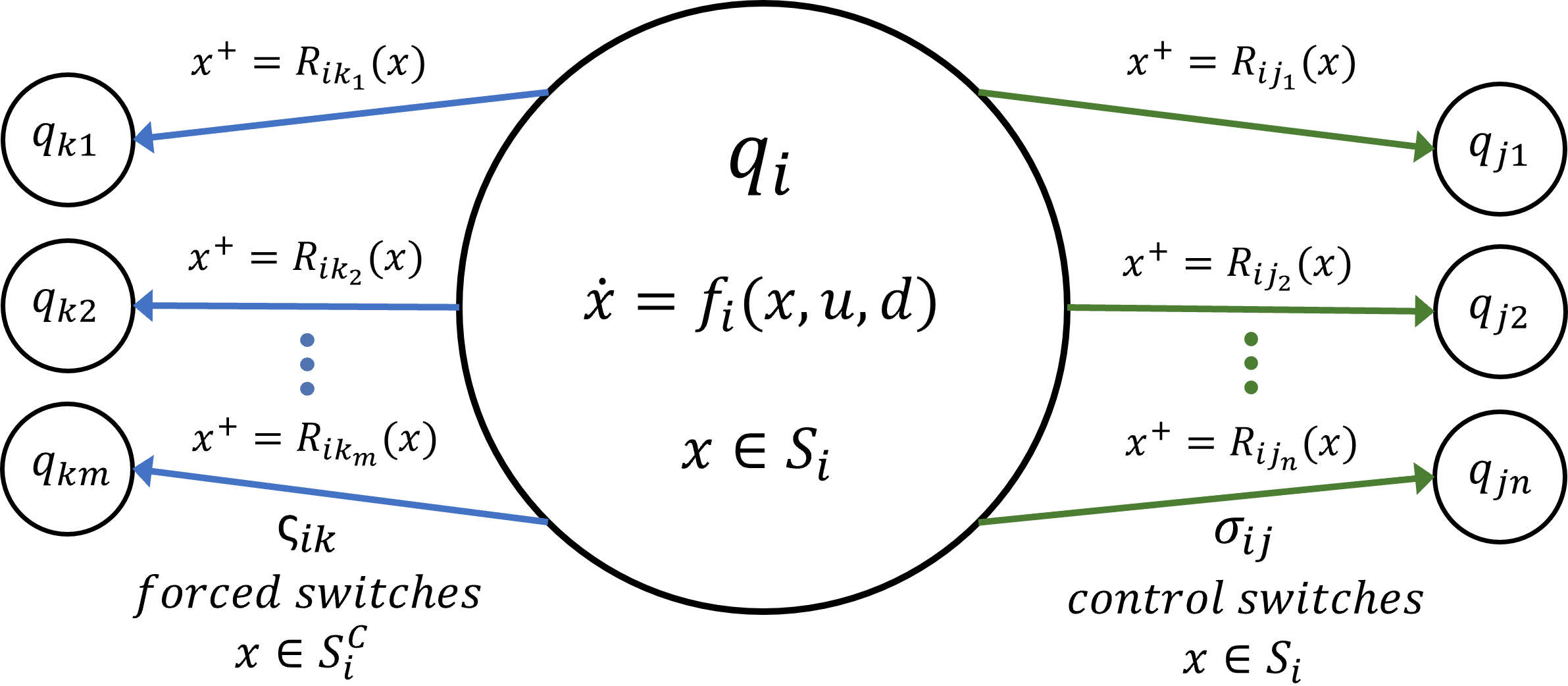}
\captionof{figure}{Hybrid dynamical system with controlled and forced transitions.}
\end{center}
\end{figure}
\hl{Let $x \in X \subset \mathbb{R}^{n_x}$ be the continuous state of the system, $u \in U \subset \mathbb{R}^{n_u}$ be the continuous control input, and $d \in D \subset \mathbb{R}^{n_d}$ be the continuous disturbance in the system.
$f: Q \times X \times U \times D \to \mathbb{R}^{n}$ defines the continuous evolution of the system for each $q \in Q$. 
$\operatorname{Inv} \subseteq (Q \times X)$ is the invariant of each discrete state, and defines the set of states for which
continuous evolution is allowed.
$\Sigma$ is a finite set of discrete actions.
$R: Q \times X \times \Sigma \to 2^{Q \times X}$ is a reset relation, which encodes the discrete transitions of the hybrid system.

For simplicity, in mode $q_i \in Q$, we denote the continuous dynamics as $f_i$ and the invariant as $S_i$.
In other words, in each discrete mode $q_i \in Q$, the continuous state evolves according to the dynamics: $\dot{x} = f_i(x, u, d)$,
with $x \in S_i \coloneq \operatorname{Inv}(q_i) \subseteq X$, $u \in U$ and $d \in D$. 
Here, $S_i$ can be thought of as the valid operation domain for mode $q_i$.} 

In mode $q_i$, the system has discrete control switches $\sigma_{ij} \in \Sigma$ that allow a \textit{controlled transition} into another discrete mode $q_{j}$, where $j \in \{j_1,j_2, \hdots,j_n\} \subset \{1, 2, \hdots, N\}$.
The controlled transition can occur only when $x \in S_i$.
\hl{Whenever the evolution of the continuous state makes $x$ exit the valid operation domain for mode $q_i$ ($x \in {S_i}^C$), the system must take one of the \textit{forced control switches}, $\varsigma_{ik} \in \Sigma$.}
This leads to a \textit{forced transition} into another discrete mode $q_{k}$ 
where $k \in \{k_1,k_2,...,k_m\} \subset \{1, 2, \hdots, N\}$. 
Note that the number of discrete modes the system can transition into may vary across different $q_i$.
Each discrete transition from $q_i$, whether controlled or forced, might lead to a state reset, which is given by the reset relation $R$ in \eqref{eq:hyb_def}.
For simplicity, we denote the state reset map upon transitioning from mode $q_i$ to $q_j$ as
$x^+ = R_{ij}(x)$, where $x$ is the state before the transition and $x^+$ is the state after the transition.

Our key objective in this work is to compute the \textbf{Backward Reachable Tube (BRT)} of this hybrid dynamical system, defined as the \hl{set of initial discrete and continuous states $(x,q)$ of the system such that starting from these states, for all disturbance inputs, there exists a control input that will eventually steer the system to the target set $\mathcal{L}$ within the time horizon $[t, T]$.} 
%
%
\hl{To mathematically define the BRT, we follow the notation in \cite{mitchell2005time}.
Let $u(\cdot)$ and $d(\cdot)$ denote the control and disturbance functions over time. 
We further assume that these functions $u(\cdot), d(\cdot)$ are drawn from the set of measurable functions:}
$$
\begin{aligned}
u(\cdot) \in \mathcal{U}(t) & =\{\phi:[t, 0] \rightarrow U: \phi(\cdot) \text { is measurable }\} \\
d(\cdot) \in \mathcal{D}(t) & =\{\phi:[t, 0] \rightarrow D: \phi(\cdot) \text { is measurable }\}
\end{aligned}
$$
\hl{Since the control and disturbance are playing against each other (a differential game in this case), it is important to address what information they know about each other's decisions. 
We will assume that the disturbance only uses 
nonanticipative strategies, that is strategies}
$$
\begin{aligned}
& \gamma \in \Gamma(t) \triangleq\{\vartheta: \mathcal{U}(t) \rightarrow \mathcal{D}(t) \mid u(r)=\hat{u}(r) \\
& \text { for almost every } r \in[t, s] \\
& \Longrightarrow \vartheta[u](r)=\vartheta[\hat{u}](r) \text { for almost every } r \in[t, s]\} .
\end{aligned}
$$
\hl{Informally, this restriction means that if disturbance cannot distinguish between control signals $u(\cdot)$ and $\hat{u}(\cdot)$ until after time $s$, then it cannot respond differently to those signals until after time $s$. Under this setting, the disturbance still has an instantaneous advantage, as it can respond after observing the current control action. 
It turns out that under nonanticipative strategies, the BRT can be obtained using a value function that is the solution of certain Hamilton-Jacobi-Isaacs PDE \cite{mitchell2005time}. 
For a more detailed discussion on non-anticipative strategies, we refer the interested readers to \cite{mitchell2005time}.}

\hl{We are now ready to formally define the BRT of the hybrid system in \eqref{eq:hyb_def}:}
\begin{equation*}
    \resizebox{1\hsize}{!}{$
\mathcal{V}(t)=\{(x_0, q_0): \forall \gamma \in \Gamma(t), \exists u(\cdot) \in \mathcal{U}(t), \sigma(\cdot), \varsigma(\cdot), \exists s \in [t, T], x(s) \in \mathcal{L}\},     
        $}
\end{equation*}
where $x(s)$ denotes the (continuous) state of the system at time $s$, starting from state $x_0$ and discrete mode $q_0$ under continuous control profile $u(\cdot)$, discrete control switch profile $\sigma(\cdot)$\footnote{If there is no discrete control being leveraged at time $s$, $\sigma(s)$ or $\varsigma(s)$ can be thought of a ``dummy'' discrete control that keeps the system in the current discrete state.}, forced control switch profile $\varsigma(\cdot)$, and a non-anticipative disturbance strategy $\gamma$.
\hl{Along with the BRT, we are also interested in obtaining the continuous and discrete control laws, i.e.,  $u(\cdot)$ and $\sigma(\cdot)$, respectively, that optimally steer the system to the target set.} 
Finally, if $\mathcal{L}$ represents the set of undesirable states for the system (such as obstacles for a navigation robot), we are interested in computing the BRT with the role of control and disturbance switched, i.e., finding the set of initial states from which getting into $\mathcal{L}$ is unavoidable, despite best control effort.
In this case, the optimal continuous and discrete control inputs ensure that the system remains outside the BRT (that is, it remains safe).

\begin{mdframed}[style=MyFrame,nobreak=false]
\textbf{Running example \textit{(Dog1D)}:} 
To illustrate our approach, we will use the following running example throughout this paper.
We consider very simplified, high-level dynamics for a robot quadruped moving in one dimension in a room with an obstacle (table). 
We present more complex hybrid dynamical systems with nonlinear dynamics in Sec. \ref{cases}.

{\centering    
\includegraphics[width=0.6\columnwidth]{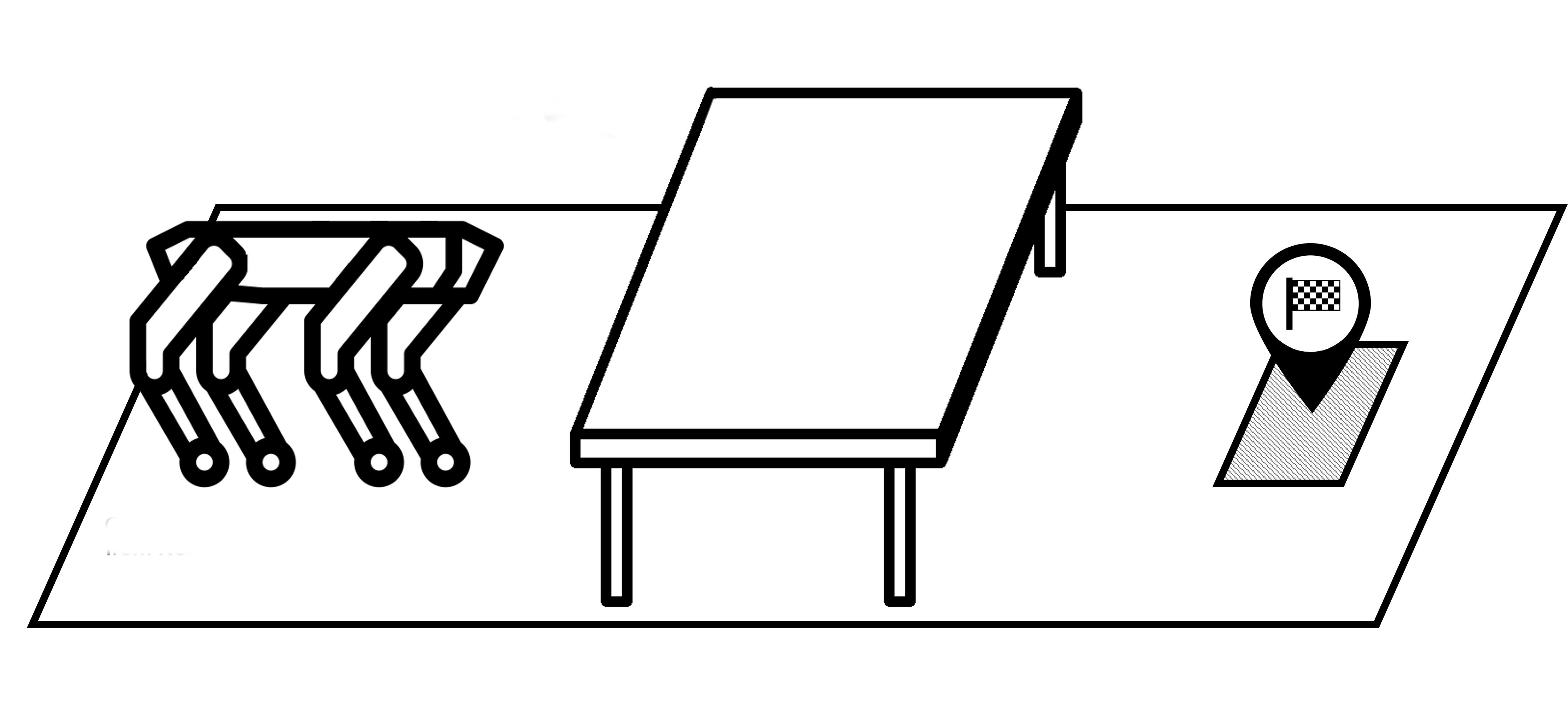}
      \captionof{figure}{A simple 1D longitudinal navigation scenario.}
      \label{fig:dog1} 
\par}

The robot's goal is to reach the marked area on the other side of the table as shown in Fig.~\ref{fig:dog1}. The hybrid system representation of this system is shown in Fig.~\ref{fig:dog1_diagram}:
 
{\centering      \includegraphics[width=0.75\columnwidth]{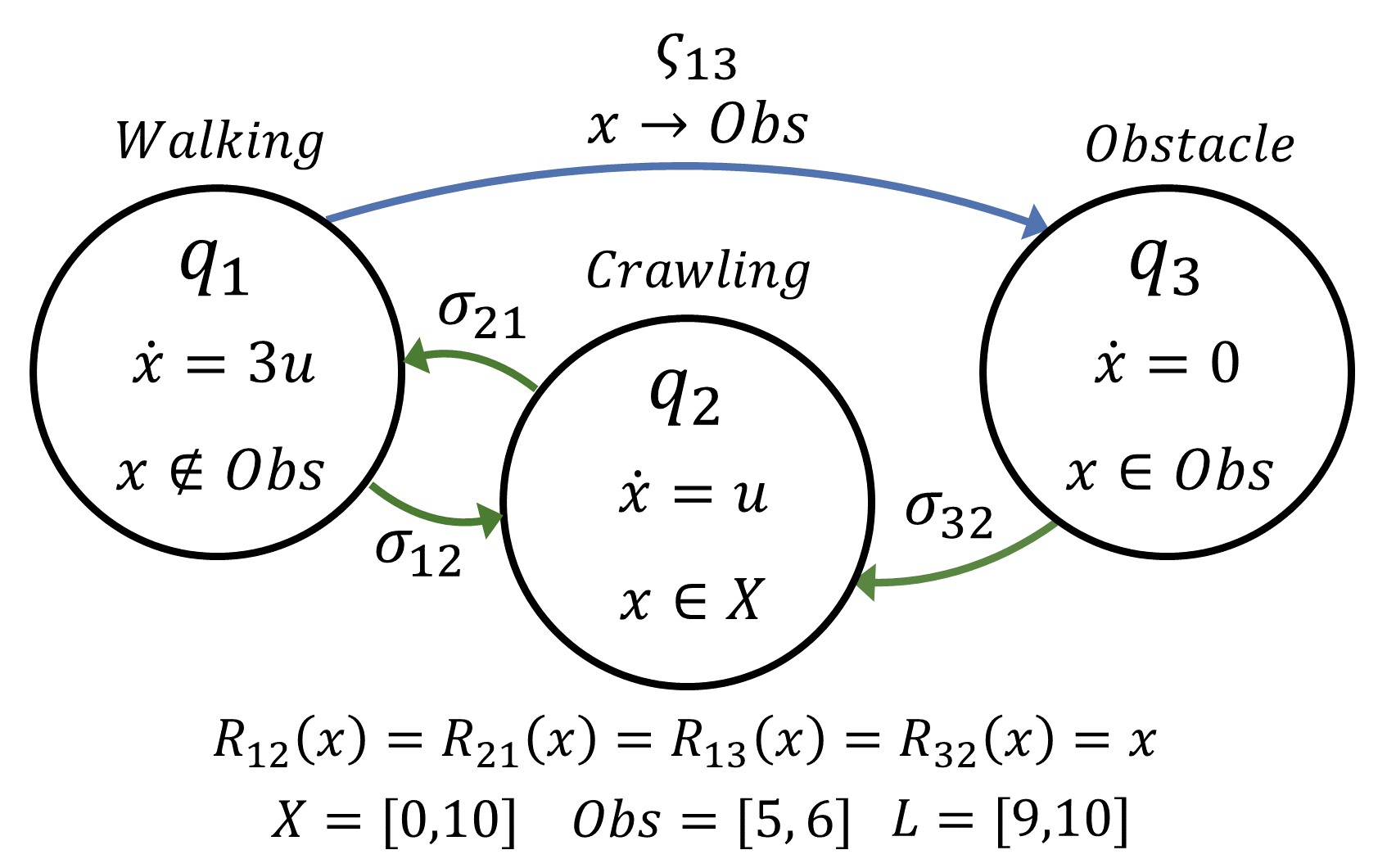}
      \captionof{figure}{Dog1D hybrid system formulation.}
      \label{fig:dog1_diagram} 
\par}
 
Here, the continuous state $x \in \mathbb{R}$ is the position of the robot and $q_i$ with $i \in \{1,2,3\}$ are the discrete modes of the system. 
$u \in [0,1]$ is the continuous control of the system.
The target set is given as the set of positions $x \in L=[9,10]$.

In this example, $q_1$ represents a walking gait; $q_2$ is a crawling gait that allows the robot to move under the table but at a slower rate compared to walking; and, $q_3$ represents frozen dynamics when the walking robot comes in contact with the table obstacle, not allowing it to pass.
The BRT for this example corresponds to all the combinations of continuous states and discrete modes $(x,q)$ from which the quadruped can reach the target set within a $5$ second time horizon.
\end{mdframed}

\section{\label{background} Background: Hamilton-Jacobi Reachability}
One way to compute the BRT for continuous-time dynamical systems is through Hamilton-Jacobi (HJ) reachability analysis. 
To illustrate HJ reachability, we assume that the system has only one discrete mode, i.e., there are no discrete transitions.

In HJ reachability, the BRT computation is formulated as a zero-sum game between control and disturbance.
This results in a robust optimal control problem that can be solved using the dynamic programming principle. 
First, a target function $l(x)$ is defined whose sub-zero level set is the target set $\mathcal{L}$, i.e. $\mathcal{L} = \{x : l(x)\leq 0\}$. The BRT seeks to find all states that could enter $\mathcal{L}$ at any point within the time horizon. This is computed by finding the minimum distance to $\mathcal{L}$ over time:
%
\begin{equation}
J(x, t, u(\cdot), d(\cdot))=\min _{\tau \in[t, T]} l\left(x(\tau)\right)
\end{equation}
Our goal is to capture this minimum distance for optimal system trajectories. Thus, we compute the optimal control that minimizes this distance (drives the system towards the target) and the worst-case disturbance signal that maximizes the distance. The value function corresponding to this robust optimal control problem is:
\begin{equation}\label{eq:hji}
 V(x, t)=\adjustlimits\sup_{\gamma \in \Gamma(t)} \inf_{u(\cdot)} \{J(x, t, u(\cdot), \gamma[u](\cdot))\},
\end{equation}
where $\Gamma(t)$ defines the set of non-anticipative strategies.
The value function in (\ref{eq:hji}) can be computed using dynamic programming, which results in the following final value Hamilton-Jacobi-Isaacs Variational Inequality (HJI-VI) \hl{\cite{bansal2017hamilton,lygeros2004reachability,mitchell2005time}}:
\begin{equation}\label{eq:pde}
\begin{array}{c}
\min \left\{D_{t} V(x, t)+\mathcal{H}(x, t), l(x)-V(x, t)\right\}=0 \\
V(x, T)=l(x)
\end{array}
\end{equation}
$D_t$ and $\nabla$ represent the time and spatial gradients of the value function. $\mathcal{H}$ is the Hamiltonian, which optimizes over the inner product between the spatial gradients of the value function and the dynamics to compute the optimal control and disturbance:
\begin{equation} \label{eqn:ham}
\mathcal{H}(x, t)=\max _{d \in D} \min _{u \in U}\nabla V(x, t) \cdot f(x, u, d) .
\end{equation}
Intuitively, the term $l(x)-V(x, t)$ in (\ref{eq:pde}) restricts system trajectories that enter and leave the target set, enforcing that any trajectory with a negative distance at any time will continue to have a negative distance for the rest of the time horizon. 
For a detailed derivation and discussion of the HJI-VI in \ref{eq:pde}, we refer the interested readers to \cite{mitchell2005time} and \cite{bansal2017hamilton}. 

Once the value function is obtained, the BRT is given as the sub-zero level set of the value function:
\begin{equation}
\mathcal{V}(t)=\{x: V(x, t) \leq 0\}
\end{equation}
The corresponding optimal control can be derived as:
\begin{equation}\label{eq:optctrl}
u^{*}(x, t)=\argmin_{u \in U} \max_{d \in D}\nabla V(x, t) \cdot f(x, u, d)
\end{equation}
The system can guarantee reaching the target set as long as it starts inside the BRT and applies the optimal control in (\ref{eq:optctrl}) at the BRT boundary. The optimal adversarial disturbance can be similarly obtained as (\ref{eq:optctrl}).

This formulation can also be used to provide safety guarantees by switching the roles of the control and disturbance in \eqref{eqn:ham}.
In that case, the BRT represents the initial states that will eventually be driven into the target by optimal disturbance, despite the best control effort.
Thus, safety can be maintained as long as the system stays outside the BRT and applies optimal control at the boundary of the BRT.

Multiple computation tools exist to compute the value function and obtain the BRT and the optimal controller.
This include methods that solve the HJI-VI in \eqref{eq:hji} numerically \cite{bansal2020provably, bui2022optimizeddp,mitchell2004toolbox,hj_reach_ASL2023} or using learning-based methods \cite{bansal2021deepreach,fisac2019bridging}.
However, one of the key limitations of HJ reachability analysis is that it assumes continuous dynamics and does not account for discrete mode switching or state resets.
Overcoming this limitation is the core focus of this work.

\begin{mdframed}[style=MyFrame,nobreak=false]
\textbf{Running example \textit{(Dog1D)}:}
We first compute the BRT for the running example for a time horizon of $T=5s$ using the classical Hamilton-Jacobi formulation with no discrete mode switching.
\hl{For the BRT computation, we use an implicit target function $l(x):=|x-9.5|-0.5$, corresponding to the goal area} in Fig.~\ref{fig:dog1}.
The obtained results are shown in Fig.~\ref{fig:dog1d_brt_bad}.
%

For the crawling gait it can be observed that the BRT (the blue shaded area) propagates through the obstacle and the BRT includes $x=4m$ because the system can move at $1m/s$ reaching the boundary of the target in exactly $5s$.
For the walking wait, even though the system is allowed to move faster, the BRT stops at the boundary of the obstacle as the walking gait gets stuck, so only states that start from the right side of the obstacle can reach the target set.

If we imagine the system being able to transition optimally between the discrete modes, it would crawl only under the table and walk elsewhere, allowing it to reach the target set starting from a walking gait and to the left of the obstacle.
As we demonstrate later, the proposed framework can reason about such optimal transitions.

{\centering   
\includegraphics[width=.90\columnwidth]{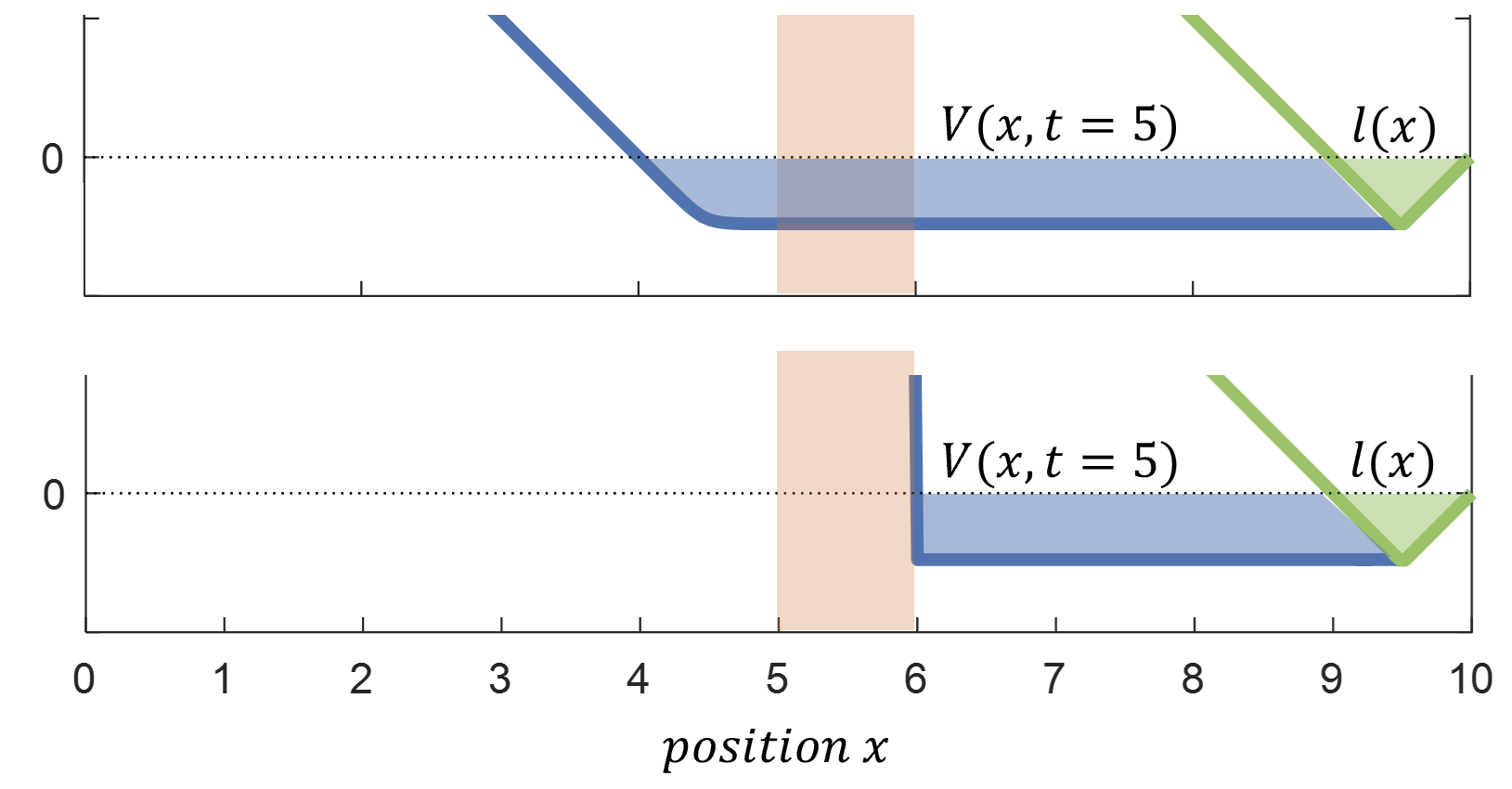}
\vspace{-0.5em}
\captionof{figure}{ (Top) Value function for the crawling gait mode. (Bottom) Value function for the walking gait mode. \hl{Both Value functions are calculated for a time horizon of $T=5s$}. The obstacle is shown in shaded red. The blue shade shows BRT as the subzero level of the value function. The green shade shows the target set as the subzero level of $l(x)$.}
\label{fig:dog1d_brt_bad} 
\par}
 
\end{mdframed}

\section{\label{approach}HJ Reachability for General Hybrid Systems}
The core contribution of this work is Theorem~\ref{theorem1}, an extension of the classical HJ reachability framework to hybrid dynamical systems with controlled and forced transitions, as well as state resets.
\begin{theorem}\label{theorem1}
\hl{Consider the hybrid dynamical system $H$ as defined in \eqref{eq:hyb_def}.
Let $l(x)$ be an implicit representation of the target set $\mathcal{L}$, i.e. $\mathcal{L} = \{x : l(x)\leq 0\}$.
Also let the value function $V(x, q_i, t)$ be the solution of the following constrained HJI-VI:} 
%

\noindent If $x \in S_i$:
%
\begin{multline*}
\min\{l(x)-V(x, q_i, t), \min_{\sigma_{ij}}V(R_{ij}(x), q_{j}, t) - V(x, q_i, t), \\D_{t}V(x, q_i, t)+ \max_{d \in D} \min_{u \in U} \nabla V(x, q_i, t)\cdot f_i(x,u,d)\} = 0,
\end{multline*}
%
%
\noindent and if $x \in S_i^C$, 
\begin{equation*}
V(x, q_i, t) = \min_{\varsigma_{ik}}V(R_{ik}(x), q_{k}, t),
\end{equation*}
%
%
with terminal time condition:
\begin{equation}
V(x, q_i, T) = l(x) 
\end{equation}
Then, the BRT for the hybrid system is given as:
%
\begin{equation}
\mathcal{V}(t)=\{(x, q): V(x, q, t) \leq 0\}
\end{equation}
\end{theorem}
Intuitively, Theorem~\ref{theorem1} updates the value function for each discrete mode $q_i$ with the optimal value across the discrete modes it can transition to. 
For states in the operation domain $S_i$, this corresponds to taking one of the control switches $\sigma_{ij}$ with their corresponding reset maps $R_{ij}(x)$ or staying in the operation mode $q_i$ itself, which will lead to a continuous flow of the value function. 
In contrast, for states outside the operation mode domain \hl{$S_i^C$}, the system has to take a forced switch $\varsigma_{ik}$ with their corresponding reset maps $R_{ik}(x)$.
Thus, the optimal value function is given by the one that takes the system to the minimum value upon the discrete transition. 
The proof of Theorem 1 formalizes this intuition using the Bellman principle of optimality and is presented in Section~\ref{proof}. 
%
Once we compute the value function, the optimal discrete and continuous controls at state $(x, q_i)$ at time $t$ are given as:
%
%
%
\begin{equation} \label{eqn:opt_control_hybrid}
\begin{aligned}
u^{*}(x, q_i, t)&=\argmin_{u \in U} \max_{d \in D}\nabla V(x, q_i, t) \cdot f_i(x, u, d) \\
\sigma_{ij^*}(x, q_i, t) &=\argmin_{\sigma_{ij}}V(R_{ij}(x), q_{j}, t) \text{~if~} x \in S_i\\
\varsigma_{ik^*}(x, q_i, t) &=\argmin_{\varsigma_{ik}}V(R_{ik}(x), q_{k}, t) \text{~if~} x \in S_i^C,
\end{aligned}
\end{equation}
where, for completeness, we add a ``dummy'' discrete control $\sigma_{ii}$ that keeps the system in the same discrete mode.
\begin{remark}
    Note that the proposed framework can easily be extended to scenarios where the forced transition mode represent an adversarial or uncertain transition instead. 
    In this case, we can use $\max$ instead of $\min$ over $\varsigma_{ik}$ in Theorem 1 to account for the worst-case behavior.
\end{remark}

\begin{remark}
    When the target set represents undesirable states of the system, we can switch the role of control and disturbance in Theorem \ref{theorem1} and during the optimal control computation in (\ref{eqn:opt_control_hybrid}) to obtain the BRT and the optimal safety controller.
\end{remark}

\noindent \textbf{Numerical implementation:} We now present an approximate numerical algorithm that can be used to calculate the value function in Theorem 1. 
It builds upon the value function calculation for the classical HJI-VI in \eqref{eq:hji}, which is solved using currently available level set methods \cite{mitchell2004toolbox}. 
Specifically, the value function is computed over a discretized state-space grid and propagated in time using a small timestep $\delta$.
After each propagation step, the value function is updated for all $(x,q)$ using Theorem~\ref{theorem1}, until the time horizon $T$ is reached. 
The detailed procedure is presented in Algorithm \ref{alg1}. 
%
\begin{algorithm}
  \caption{Value function computation for hybrid dynamical system}\label{alg1} 
 \textbf{Input:} $l(x), T, \delta$\\
 \textbf{Output:} $V(x,q_i,t) \forall i$\\
  \textbf{Initialization:} $V(x,q_i,T)=l(x) \forall i$; \quad $t=T$\\
\While{$(t > 0)$}{
\ForEach{$(q_i \in Q)$}{
$V(x,q_i,t-\delta)=V(x,q_i,t) +
    \newline \hspace*{5em}\displaystyle \max_{d} \min_{u}\nabla V(x, q_i, t)\cdot f_i(x, u, d) \delta$\\
}
\ForEach{$(q_i \in Q)$}{
\uIf{$(x \in S_i$)}{
    $\displaystyle V(x, q_i, t-\delta) =\min\{ l(x), V(x, q_i, t-\delta),
    \newline \hspace*{6.8em} \min_{\sigma_{ij}}V(R_{ij}(x), q_{j}, t-\delta) \}$
  }
  \uElseIf{$(x \in S_i^C$))}{
     $\displaystyle V(x, q_i, t-\delta) =
 \min_{\varsigma_{ik}}V(R_{ik}(x), q_{k}, t-\delta) $
  }
}
$t=t-\delta$
}
\end{algorithm}
%

It is important to stress the remarkable similarity of this new hybrid reachability algorithm to its continuous counterpart in \eqref{eq:hji}. 
Indeed, there are only two key differences: (1) we now propagate $N$ value functions simultaneously (corresponding to $N$ discrete modes), as opposed to just one value function.
(2) We ``adjust'' the value function for each discrete mode to account for discrete switches (the second for-loop in Algorithm \ref{alg1}).
\hl{Despite this simplicity, the algorithm simultaneously reasons about all possible discrete and continuous transitions to optimally steer the system to the target set. In the following sections, we show how this joint reasoning generates optimal yet intuitive behaviors in our simulation and hardware experiments, like a jumping robot crouching to build momentum in preparation for a jump transition or a quadruped opting for a slower gait to take a sharp corner.} 

\hl{\textit{Complexity analysis of Algorithm \ref{alg1}}. Algorithm \ref{alg1} computes the BRT over a state-space grid, using a finite time step.  
Assuming there are $M$ grid points per state dimension, we have a total of $M^{n_x}$ grid points, where $n_x$ is the number of continuous states. 
At each timestep, for each discrete mode, the computation of the value function requires evaluating all potential discrete transitions from the current mode (the inner for loop in Algorithm \ref{alg1}). This results in $\mathcal{O}(N \cdot M^{n_x})$ computations per timestep, per mode. Since there are N discrete modes, the computational requirement per timestep scales to  $\mathcal{O}(N^2 \cdot M^{n_x})$. Finally, the total number of timesteps is proportional to the time horizon $T/\delta$, resulting in a total complexity for Algorithm 1 of  $\mathcal{O}(T \cdot N^2 \cdot M^{n_x} / \delta)$. 
In summary, the computational complexity of Algorithm 1 scales linearly with the time horizon, quadratically with the number of discrete modes, and exponentially with the number of continuous state dimensions.
}

\begin{remark}
\hl{Note that Algorithm \ref{alg1} to compute the value function is approximate. Specifically, Algorithm 1 might potentially require infinite grid resolution and infinitesimally small time steps to compute the value function exactly. Hence, the algorithm may not terminate in finite time for general hybrid systems. This limitation echoes the broader undecidability challenges in computing exact reachable tubes for hybrid systems \cite{lygeros1998controller, henzinger1995s}. 
Nevertheless, the proposed algorithm has proven to be quite effective in practice for approximating BRTs, as we demonstrate later by several case studies and real robot experiments.}
\end{remark}
\section{\label{proof}Theoretical Result: Proof of Theorem 1}
%
In this section, we present a proof for Theorem \ref{theorem1}.
We start with an analogous formulation to HJ reachability in Section \ref{background} with a target function $l(x)$ such that the target set is defined as, $\mathcal{L} = \{(x) : l(x)\leq 0\}$. 
The BRT seeks to find all states that could enter $\mathcal{L}$ at any point within the time horizon. This is computed by finding the minimum distance to $\mathcal{L}$ over time.
%
%

 The value function is defined as the cost incurred under optimal discrete and continuous controls that minimizes distance to the target and optimal disturbances that maximizes it. 
 We first consider the case when the state belongs to the interior of the domain $S_i$.
 The value function is given as:
\vspace{-0.2em}
\begin{equation}\label{eq:hyb_value_def}
V(x, q_i, t) = \max_{\gamma \in \Gamma} \min_{u \in \mathcal{U}} \min_{\sigma_{ij}}\min_{s \in[t, T]} l(x(s))
\end{equation}
With terminal time condition given by:
\vspace{-0.2em}
\begin{equation}
V(x, q_i, T) = l(x) 
\end{equation}

The optimal decision making for discrete transitions is pictorially shown in the figure below.  
\begin{figure}[h!] 
\begin{center} 
\vspace{-0.5em}
\includegraphics[width=0.85\columnwidth]{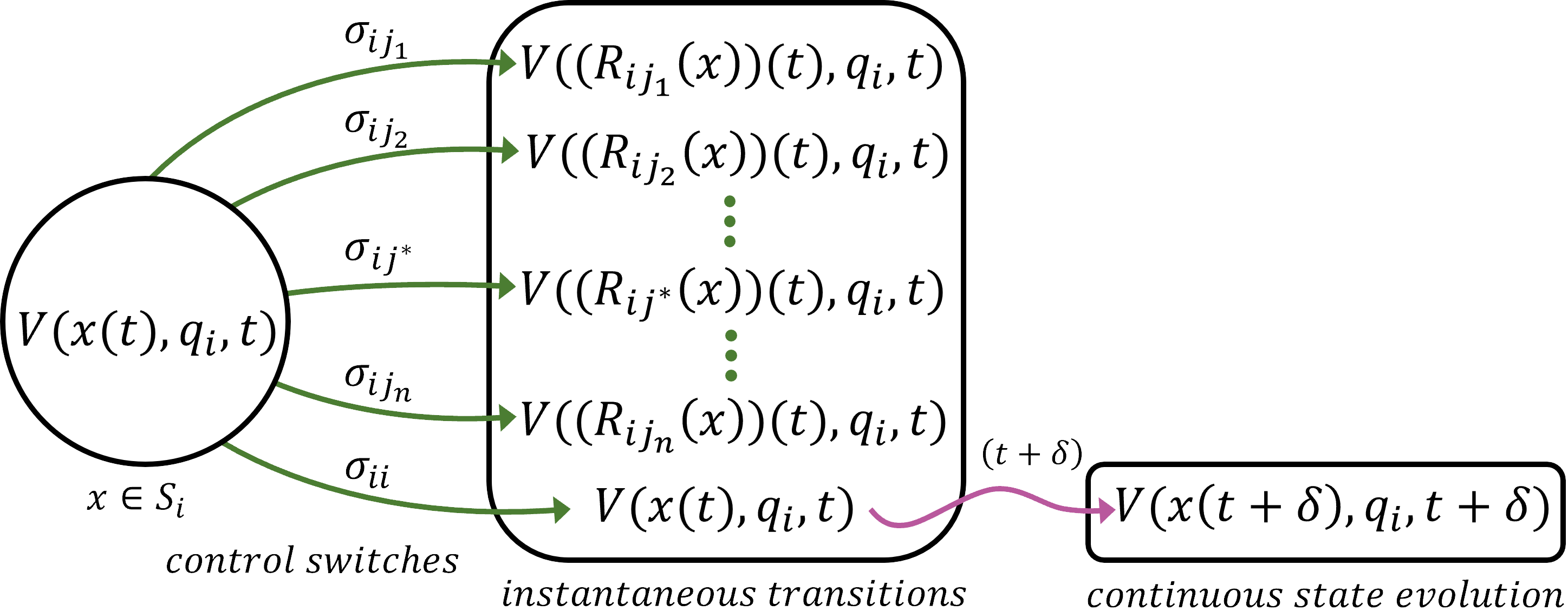}
\vspace{-1.em}
\end{center}
\end{figure}
The discrete transitions are shown in green and continuous control flow in purple.
We also add a ``dummy'' discrete transition, $\sigma_{ii}$, that keeps the system in the current discrete mode.
As per \eqref{eq:hyb_value_def},
the overall value function is given by the minimum cost across all these possible trajectories of the system.
From here, we consider two different cases:

\noindent \textbf{Case 1:} System switches to a new discrete mode $j^*$, continuous state immediately transitions to $R_{ij^*}$. 
In this case, the optimal cost incurred by the system from the new state $(R_{ij^*}(x), q_{j^*})$ is, by definition, given as $V(R_{ij^*}(x), q_{j^*}, t)$. 
Thus, the optimal cost across all possible discrete transitions is given as:
%
\vspace{-0.2em}
\begin{equation}
V(x, q_i, t) =\min_{\sigma_{ij}}V(R_{ij}(x), q_{j}, t)
\end{equation}

\noindent \textbf{Case 2:} System evolves in mode $q_i$, i.e., it takes the dummy switch $\sigma_{ii}$. 
In this case, for a small time step $\delta$ and using the dynamic programming principle for the cost function in \eqref{eq:hyb_value_def}, we have:
%
\vspace{-0.2em}
\begin{align} \label{eqn:cont_flow}
V(x, q_i, t) & = \max_{\gamma \in \Gamma} \min_{u \in \mathcal{U}} \min\{\min_{s \in[t, t+\delta]} l(x(s)), \nonumber \\ 
& \quad \quad \quad \quad \quad \quad \quad \quad
 V\left(x(t+\delta), q_i, t+\delta\right)\} \nonumber\\
& \approx \max_{\gamma \in \Gamma} \min_{u \in \mathcal{U}} \min\{l(x(t)), V\left(x(t+\delta), q_i, t+\delta\right)\}
\end{align}

Approximating the value function at the next time step with a first order Taylor expansion:
\begin{align}
V\left(x(t+\delta), q_i, t+\delta\right) \approx & V(x, q_i, t) + \nonumber \\
& D_{t}V(x, q_i, t)\delta+\nabla V(x, q_i, t)\cdot\delta x,
\end{align}
where the change in the state $\delta x$ can be approximated as $f_i(x,u,d) \delta$. Ignoring the higher order terms and plugging the Taylor expansion in \eqref{eqn:cont_flow}:
%
\vspace{-0.2em}
\begin{multline}
V(x, q_i, t) = \min\{l(x(t)),\\ V(x, q_i, t) + D_{t}V(x, q_i, t)\delta+\\ \max_{d \in D} \min_{u \in U} \nabla V(x, q_i, t)\cdot f_i(x,u,d) \delta \} 
\end{multline}
%

\noindent \textbf{Combining Cases:} The optimal value function is given by the minimum across the two cases:
\begin{multline}
V(x, q_i, t) = \min\{l(x),\\ V(x, q_i, t) + D_{t}V(x, q_i, t)\delta+\\ \max_{d \in D} \min_{u \in U} \nabla V(x, q_i, t)\cdot f_i(x,u,d) \delta ,\\ \min_{\sigma_{ij}}V(R_{ij}(x), q_{ij}, t)\} 
\end{multline}
Subtracting the value function $V(x, q_i, t)$ from both sides:
\begin{multline}
0 = \min\{l(x(t))-V(x, q_i, t),\\\delta (D_{t}V(x, q_i, t)+ \max_{d \in D} \min_{u \in U} \nabla V(x, q_i, t)\cdot f_i(x,u,d)) ,\\ \min_{\sigma_{ij}}V(R_{ij}(x), q_{ij}, t) - V(x, q_i, t) \} 
\end{multline}
Since the above statement holds for all $\delta>0$, we must have: 
\begin{multline}
0 = \min\{l(x(t))-V(x, q_i, t),\\D_{t}V(x, q_i, t)+ \max_{d \in D} \min_{u \in U} \nabla V(x, q_i, t)\cdot f_i(x,u,d) ,\\ \min_{\sigma_{ij}}V(R_{ij}(x), q_{j}, t) - V(x, q_i, t) \} 
\end{multline}
\hl{Note that the procedure presented to derive the HJI-VI for continuous evolution of the value function is an informal proof based on the Taylor series expansion that assumes the value function to be differentiable, which may not be true, and only the existence of a viscosity solution can be ensured \cite{mitchell2005time, lygeros2004reachability}.
Nevertheless, this HJI-VI is known to hold even when the value function is non-differentiable, formal proof can be found in \cite{lygeros2004reachability}.
We omitted a detailed derivation for brevity purposes.} 

For states outside the domain of the discrete mode $q_i$ (i.e., $x \in S_i^C$), our discrete decision is over the set of available forced switches $\varsigma_{ik}$. For this case the proof is analogous to the Case 1 previously considered, but without the option to stay in $q_i$. The value function is then given by:   
\vspace{-0.2em}
\begin{equation}
V(x, q_i, t) =\min_{\varsigma_{ik}}V(R_{ik}(x), q_{k}, t)
\end{equation}
\section{\label{cases}Case Studies}
\subsection{\label{case_air3d}Running example (Dog1D)}
We now apply the proposed method to compute the BRT for the running example.
To implement Algorithm 1, we build upon helperOC library and the level set toolbox~\cite{mitchell2004toolbox}, both of which are used to solve classical, continuous-time HJI-VI.

We use a grid of 301 points over the $x$ dimension for each of the 3 discrete operation modes. 
The overall calculation takes 12.125s running on an Intel Core i5-6200U CPU @ 2.30GHz, this hardware is used across all simulation studies. 
Slices of the BRT starting in mode $q_1$ (the walking gait) for different time horizons are shown in Fig.~\ref{fig:dog1d_brt}:
\begin{figure}[h!] 
\begin{center} 
\vspace{-.5em}
\includegraphics[width=0.85\columnwidth]{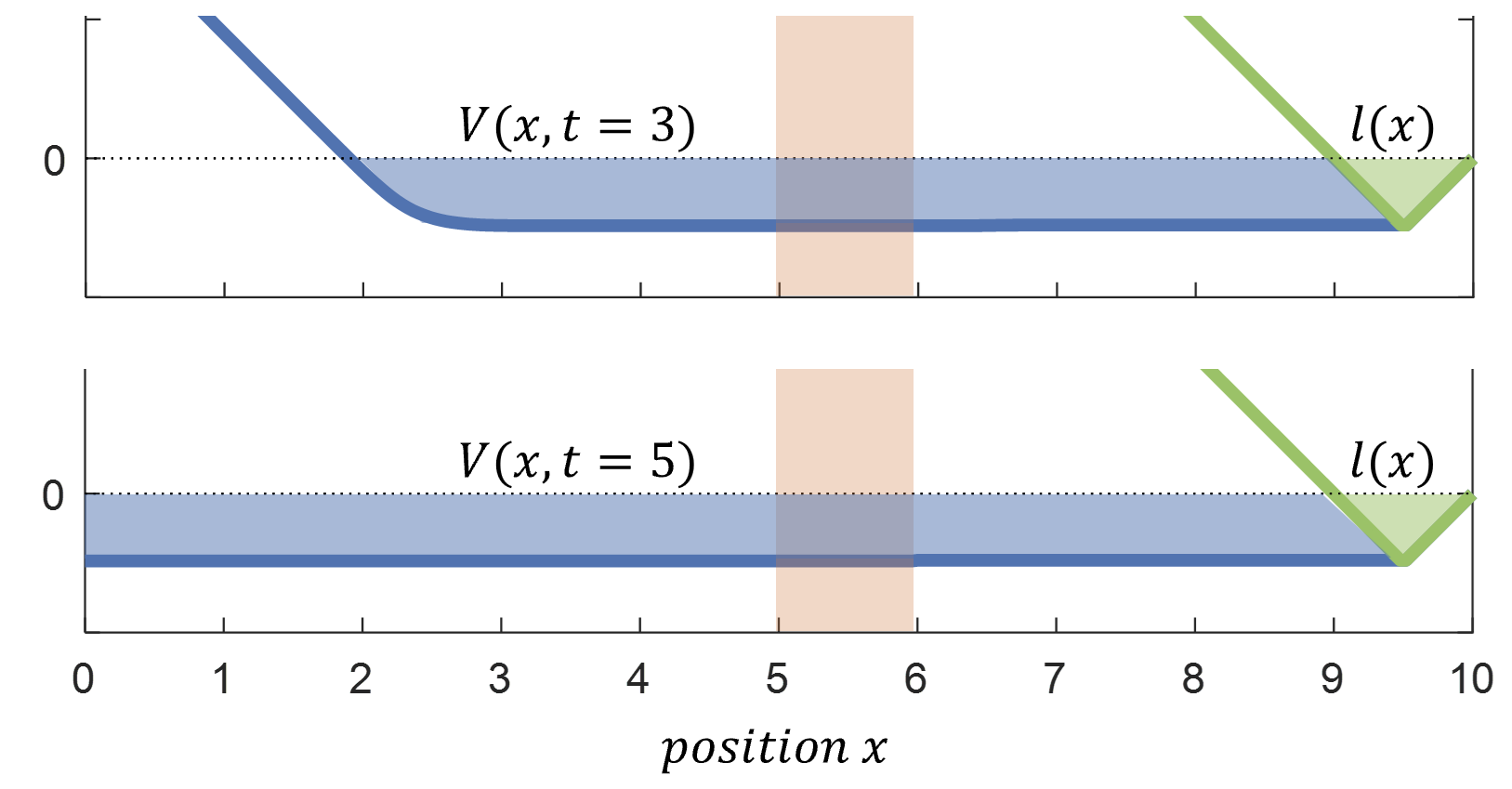} 
\vspace{-0.5em}
\caption{Value functions for states starting in a walking gait for a $3$ and $5$ seconds time horizon. 
Area within the obstacle is shown in shaded red. Blue shade shows BRT as the subzero level of the value function. Green shade shows the target set as the subzero level of the implicit target function $l(x)$.}
\vspace{-1.0em}
\label{fig:dog1d_brt} 
\end{center}
\end{figure}
The intuitive solution for the optimal decision-making in this scenario is to use the walking gait everywhere except in the obstacle area. 
The fast walking gait allows the system to reach the target quicker, while crawling allows to expand the reachable states beyond the obstacle. 
The computed solution using our algorithm indeed aligns with this intuition.
For example, if we consider the top value function for the 3-second time horizon in Fig. \ref{fig:dog1d_brt}, we can observe that the limit of the BRT is $7m$ away from the target set, which coincides with the intuitive solution of walking for $1s$ covering $3m$, then crawling under the table for $1s$ covering its $1m$ width, to finally go back to walking for $1s$ covering other $3m$.

The bottom value function in Fig. \ref{fig:dog1d_brt} shows the BRT corresponding to a $5s$ time horizon.
Compared to Fig. \ref{fig:dog1d_brt_bad}, where the BRT stopped growing beyond the obstacle boundary, we can see how the proposed algorithm can optimally leverage discrete transitions along with continuous control to reach the target set from a wider set of initial states. Additionally, even though we are calculating the reach BRT for the target set, we get an obstacle avoidance behavior, as any state that reaches the target should never touch the obstacle as it freezes the dynamics and halts the robot.


\subsection{\label{case_jump}One-Legged Jumper}
\begin{figure}[h!] 
\begin{center} 
\vspace{-.5em}
\includegraphics[width=.65\columnwidth]{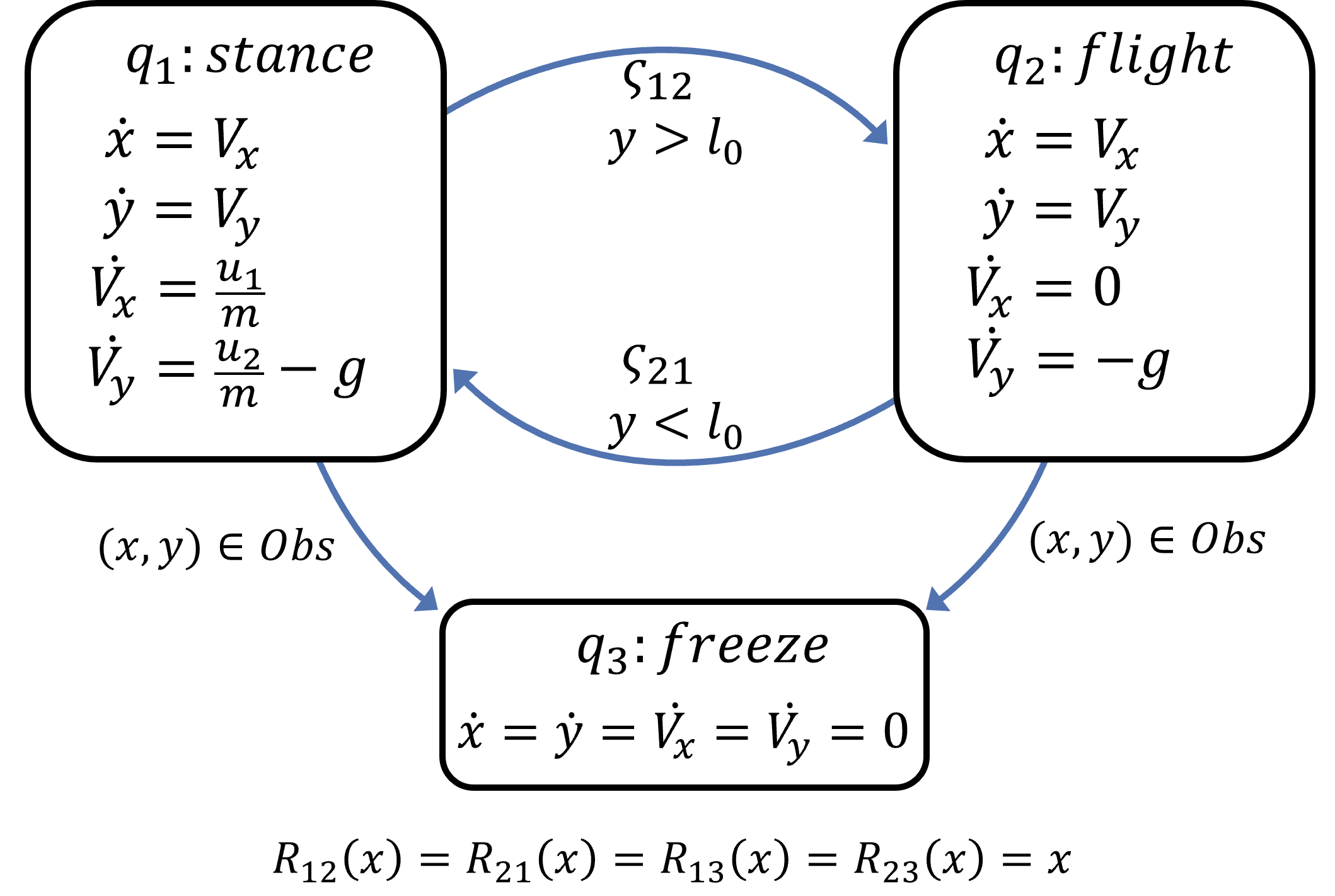} 
\vspace{-.0em}
\caption{Hybrid formulation for the planar jumping robot.}
\vspace{-1.0em}
\label{fig:dog_jump_diag} 
\end{center}
\end{figure}

\begin{figure}[h!] 
\begin{center} 
\vspace{-.5em}
\includegraphics[width=.95\columnwidth]{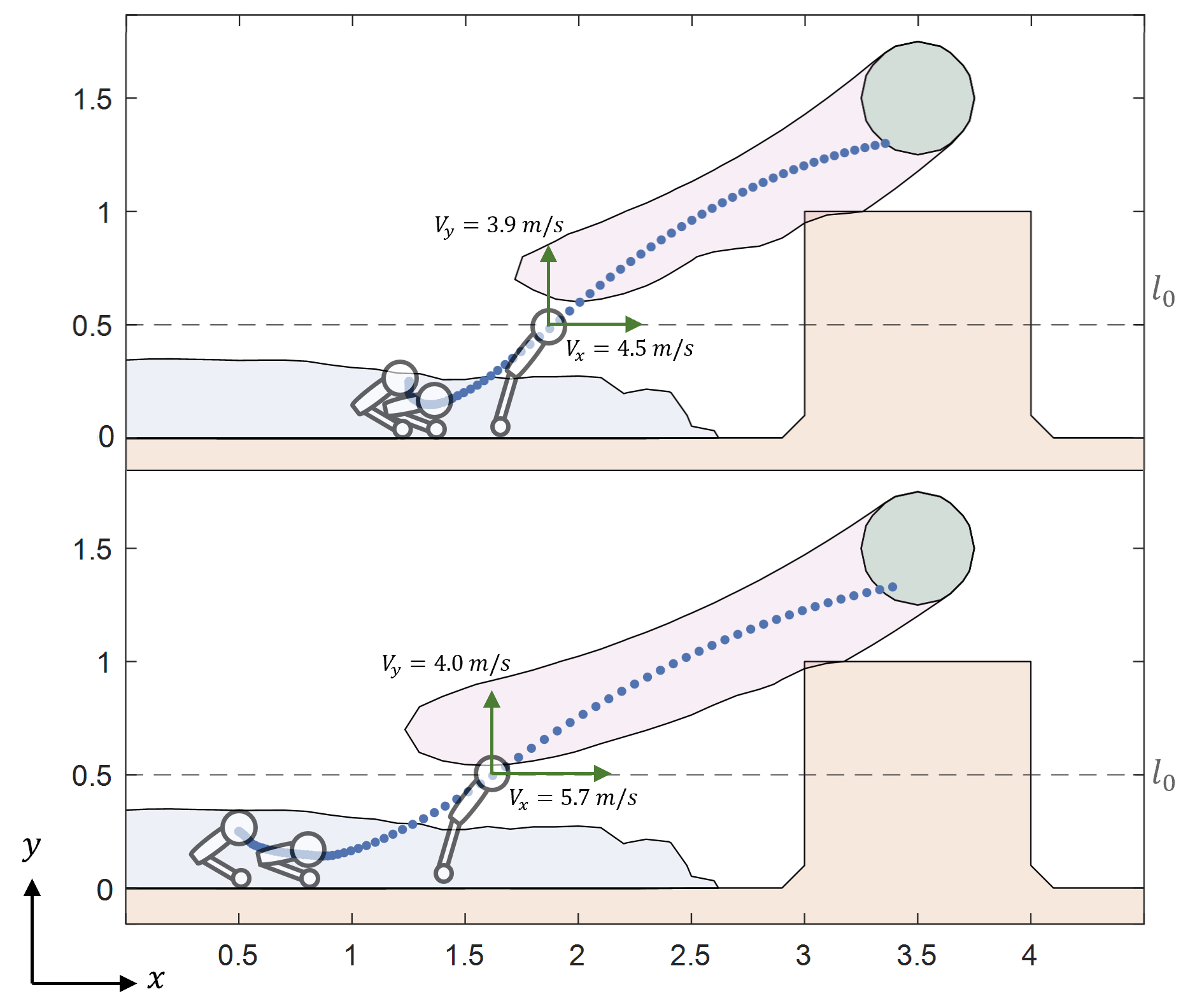} 
\vspace{-.0em}
\caption{BRT and trajectories for planar jumping robot.}
\vspace{-1.0em}
\label{fig:dog_jump_traj} 
\end{center}
\end{figure}

As a more challenging case study, we now consider a planar one-legged jumping robot that wants to reach a goal area on top of a raised platform (Fig. \ref{fig:dog_jump_traj}).
The center of mass dynamics of the jumper robot are shown in Fig~\ref{fig:dog_jump_diag}.
The robot has two main operation modes: a stance mode and a flight mode. 
The stance mode is active when the leg is in contact with the ground, which allows the leg to exert force and accelerate in xy directions.
The flight mode is activated when the leg is no longer in contact with the ground, which is modeled as an unactuated ballistic flight. 
Additionally, we consider a freeze mode that will permanently halt the dynamics if the center of mass of the robot collides with the terrain; this modeling allows us to obtain the characteristic of a reach-avoid scenario while only doing reach calculations, since any trajectory that goes through the terrain will be frozen and excluded of the reach BRT as it will never reach the target set.

In this system, the continuous state is given by $[x,y,V_x,V_y]^T$, representing both positions and velocities in the $xy$ plane. Here $u_1$ and $u_2$ are the force inputs in the $x$ and $y$ directions, both with an actuation range of $0-30 N$.
$l_0=0.5m$ is the length of the robot leg,  $m=1kg$ is the mass of the robot, and $g$ is the acceleration due to gravity. 
The target set is a circular area in the $xy$ plane on top of a raised platform, which is considered to be part of the terrain along with the ground; the objective here is to find all the initial states of the robot in the stance mode from which a successful jump into the target set is guaranteed without colliding with the platform.

\hl{Fig.~\ref{fig:dog_jump_traj} shows the obtained BRT using the proposed method over a grid with $[51, 31, 21, 21]$ points for a time horizon of $T = 0.8s$. 
The BRT calculation took 7.5 minutes using the proposed approach.}
The terrain is shown in light brown and the target set is shown in green.
The blue area represents the slice of the BRT when the robot starts at rest in the stance mode, i.e., all the initial stance positions of the robot from which it is eventually guaranteed to reach the target set under the optimal controller obtained through the proposed method. 

We also show the robot trajectories to the target set under the obtained optimal controller from two such initial configurations. 
The optimal policy shows fascinating yet intuitive results in this study case.
In both trajectories, we can observe how the robot first crouches to build upon enough vertical and horizontal momentum in preparation for a transition into the flying mode. 
We can also observe how the optimal policy leads to a longer standing trajectory for an initial condition that is further away from the target set.
This is because the robot needs to build up sufficient horizontal velocity for a longer flying trajectory. 
This qualitative analysis was corroborated by the speed of the robot just before transitioning to the flight mode.
For the top plot, the robot launch speed is $(V_x,V_y) = (4.5,3.9) m/s$, while it is $(V_x,V_y) = (5.7,4.0) m/s$ for the bottom plot, showing that both trajectories have similar vertical takeoff velocities but the state that is farther takes off with a larger horizontal velocity.

We show the robot BRT in the flight mode at the launch speeds in red.
In other words, the red region represents all initial positions of the robot from which it can reach the target set under the flight mode with the indicated launch speed.
Thus, the proposed approach allows us to reason about the continuous control profile that can then enable a successful discrete transition from the blue to the red BRT in order to eventually reach the target set.
This simultaneous reasoning of continuous and discrete transitions enable the proposed framework to ensure liveness from a broader set of initial configurations.

\hl{As a baseline, we also used the iterative reachability algorithm proposed in~\cite{air_3modes_200l} to compute the BRT for the one-legged jumper system.
It is worth noting that this baseline algorithm doesn't directly support the computation of finite time reachable sets. Here, we modify the algorithm in~\cite{air_3modes_200l} to compute the continuous reach-avoid operator for $T=0.8s$ in each mode in each iteration. 
This makes sure that we get a conservative approximation of the reach set. 
The BRT computation required two iterations to include the initial states shown in Fig. \ref{fig:dog_jump_traj} with the BRT. 
The computation took a total of 15.6 minutes, resulting in a 2x efficiency in the BRT computation under the proposed framework.
}

\subsection{\label{case_plane}Two-Aircraft Conflict Resolution}
We next consider the two-aircraft conflict resolution example presented in \cite{air_3modes_200l}.
Here, two aircraft flying at a fixed altitude are considered. 
Each aircraft controls its speed; we assume that the first aircraft is controllable, while the other aircraft's speed is uncertain (within a known interval) and hence modeled as an adversarial disturbance to ensure safety. 
Using relative coordinates, the continuous dynamics of the system are described by:
\begin{equation}
\begin{aligned}
\dot{x}_r & =-u+d \cos \psi_r+\omega_u y_r \\
\dot{y}_r & =d \sin \psi_r-\omega_u x_r \\
\dot{\psi}_r & =\omega_d-\omega_u,
\end{aligned}
\end{equation}
%
%
%
where state $[x_r,y_r,\varphi_r]^T$ is the relative $xy$ position and heading between the aircraft.
$u$ and $d$ are the velocities of the two aircraft, and $\omega_u$ and $\omega_d$ are their turning rates.

\begin{figure}[h!] 
\begin{center} 
\vspace{-.5em}
\includegraphics[width=.95\columnwidth]{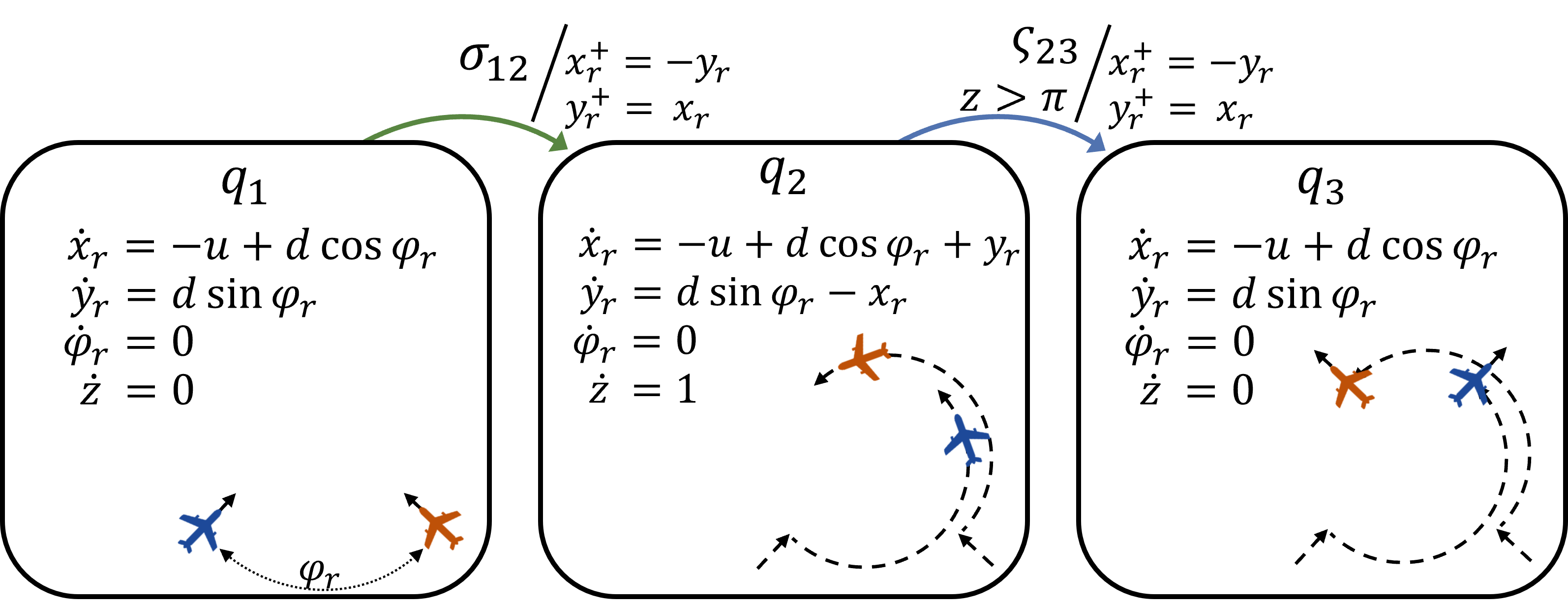} 
\vspace{-.0em}
\caption{Hybrid formulation for the two-aircraft conflict resolution protocol.}
\vspace{-2.0em}
\label{fig:air_diag} 
\end{center}
\end{figure}
The conflict resolution protocol consists of three different operation modes, shown in Fig. \ref{fig:air_diag}. 
The aircrafts begin in mode $q_1$ with a straight flight ($\omega_u=\omega_d=0$), keeping a constant relative heading. 
In this mode, the aircrafts are on a collision course.
At some time, the aircraft can begin the collision avoidance maneuver by taking a controlled switch $\sigma_{12}$, transitioning the system into mode $q_2$ and an instantaneous heading change of $\pi/2$.
In mode $q_2$, the aircraft undergo a circular flight path, where both aircraft use the same constant turning rate ($\omega_u=\omega_d=1$). Once the aircraft undergo a heading change of $\pi$ radians, the aircraft are forced to switch to mode $q_3$, making another instantaneous $\pi/2$ turn and resuming their straight flight ($\omega_u=\omega_d=0$). 
The two aircraft are now on a collision-free path.
To keep track of the transition time on configuration $q_2$ we add an additional timer state $z$. 

Unlike previous case studies, here we are interested in the safety problem, i.e., we wanted to compute an avoid BRT -- the set of all starting states from which a collision is inevitable between the vehicles, despite the best control effort. 
Consequently, we switch the role of control and disturbances in Theorem \ref{theorem1}.
Specifically, we are interested in finding the set of all initial states in mode $q_1$ from which a collision cannot be averted between the two aircrafts under the above protocol, despite the best control effort. 
Thus, the complement of the BRT will represent the safe states for the system.

For BRT calculations, we consider a circular collision target set of 5 units around the controlled aircraft.
This corresponds to the two aircraft being in close proximity of each other, also referred to as ``loss of separation''.
We use an initial relative heading of $\varphi_r=2\pi/3$ (shown in Fig. \ref{fig:air_diag}) and a time horizon of $T=5s$.
For aircraft velocities we consider $u \in [1.5,3] m/s$ and $d \in [2,4] m/s$. 

The BRT computation for this conflict resolution problem has been particularly challenging for a few different reasons: (a) the presence of nonlinear dynamics; (b) the presence of a controlled switch in $q_1$; (c) discontinuous state resets upon transitions between modes; and (d) uncertainty in the speed of the second aircraft.
\hl{An algorithm to compute the BRT for this problem has been proposed in \cite{air_3modes_200l}, which we use as a baseline for this example. 
It uses an iterative reachability algorithm, wherein the BRT is iteratively refined in each discrete mode (using a continuous reach-avoid operator) based on the last computed BRT and the discrete predecessor maps that are hand-coded.
Instead, we use the HJI-VI in Theorem \ref{theorem1} to compute the BRT in one shot without any explicit hand-coding of discrete transitions. 
The calculations are carried on a $[x_r,y_r,z]$ grid of $[301,301,11]$ points giving a spatial resolution of $\delta x = 0.1$.
$\varphi_r$ has null dynamics in all operation modes and is not considered in the grid.}

\hl{The algorithm in \cite{air_3modes_200l} takes three iterations to converge, with a total computation time of 16 minutes.
In contrast, our proposed method took 3.9 minutes to obtain the same BRT, demonstrating a 4x increase in computational efficiency.
The two BRTs are very similar, disagreeing only on $0.043\%$ of the points defined by the calculation grid, which are attributed to the numerical issues.}

\begin{figure}[h!] 
\begin{center} 
\includegraphics[width=0.95\columnwidth]{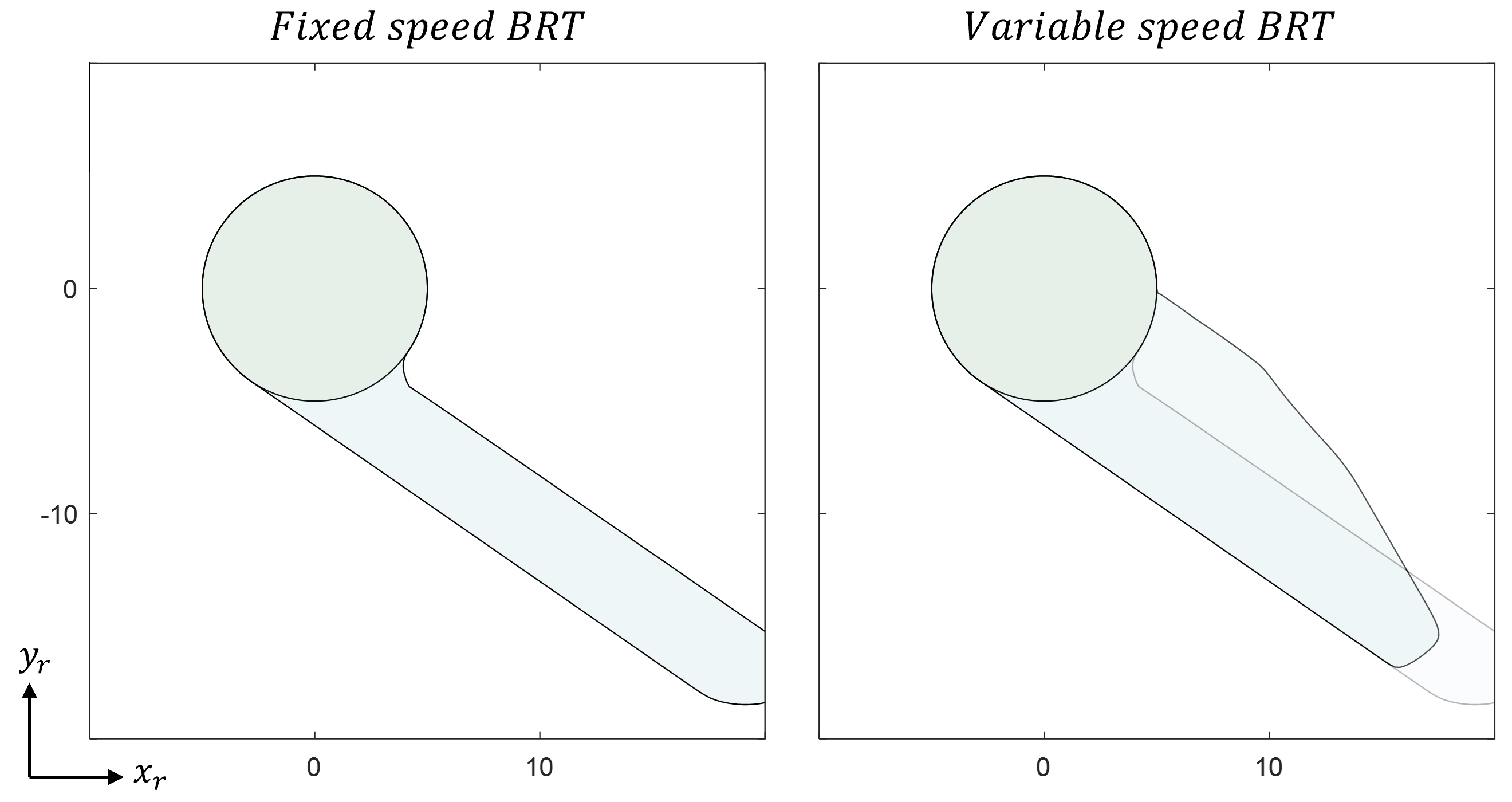} 
\caption{Backward reachable tube for two-aircraft conflict resolution starting from mode $q_1$. (Left) BRT for fixed speed of aircraft. (Right) BRT for aircraft under optimal control and disturbance.}
\vspace{-1em}
\label{fig:air_brt} 
\end{center}
\end{figure}

The obtained BRTs starting from the operation mode $q_1$ are shown in Fig.\ref{fig:air_brt}. 
They can be interpreted as follows: if the relative coordinates of the aircraft belong to a state outside the BRT, the collision avoidance maneuver can be initiated while ensuring a collision can always be avoided as long as the optimal discrete and continuous control is applied.
On the other hand, if the relative state starts inside the blue region, a collision cannot be averted under the proposed protocol.

The left BRT matches with the results of the case presented in \cite{air_3modes_200l} where both aircraft keep their speed constant at their maximum value. 
Intuitively, the interior of the BRT corresponds to states where the adversarial airplane can place itself behind the controlled aircraft and use its higher velocity to force a collision. 

Furthermore, the right BRT considers the scenario where, in addition to the control on the transition, both aircraft control their velocities optimally to avoid/force a collision with the other aircraft.
The changes in the BRT are reflected in the shaded areas, where the growth in the BRT closer to the target set corresponds to cases where the adversarial aircraft slows down to align itself with the controlled aircraft and then use its higher velocity to force a collision. 
The decrease in size near the tail corresponds to states where the blue airplane slows down to avoid a collision. 
%

%
\section{\label{hw_exp}Hardware Experiments}
\begin{figure}[H]
\begin{center} 
\vspace{-1.0em}
\includegraphics[width=0.95\columnwidth]{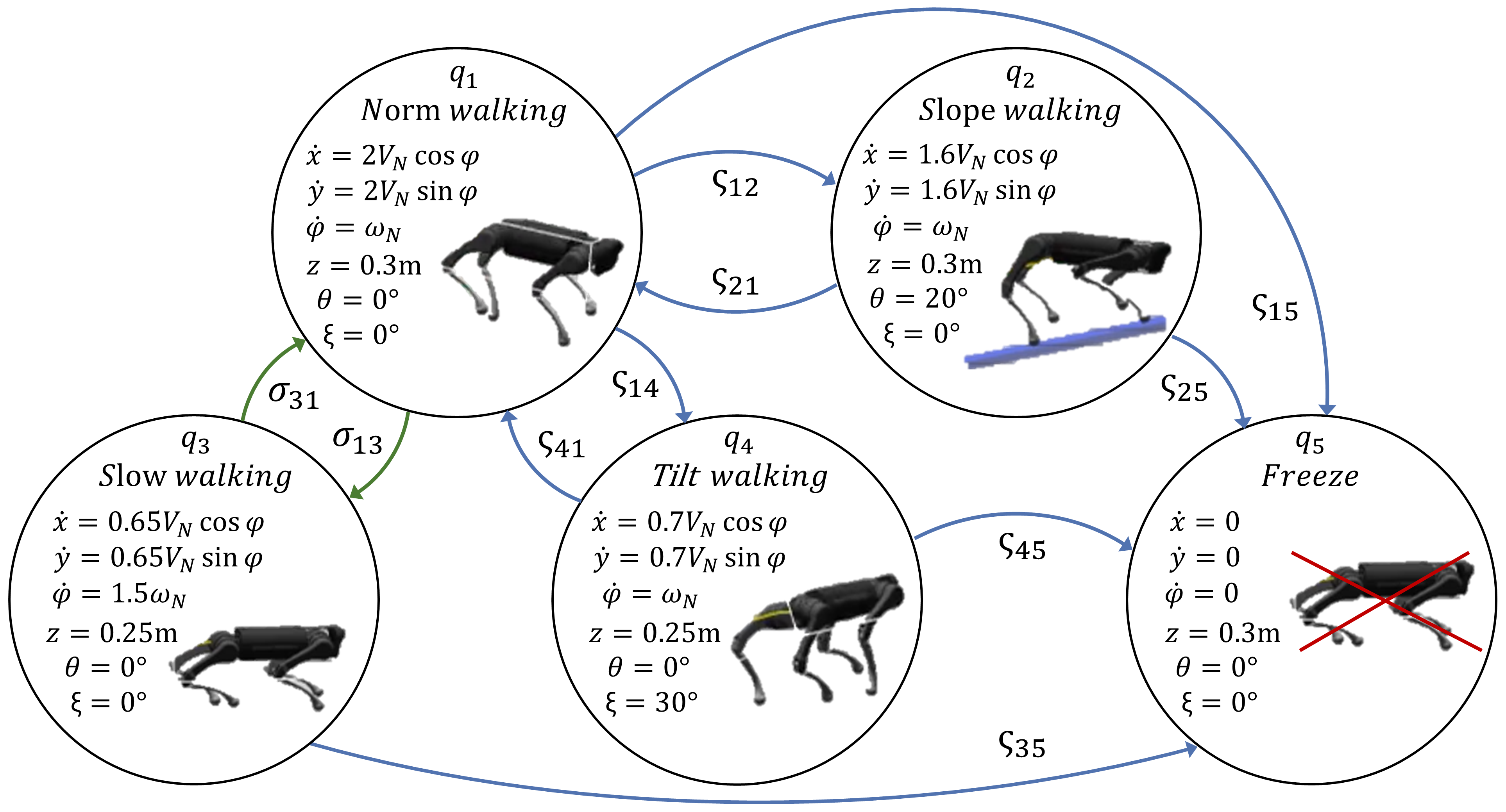} 
\caption{Hybrid control modes for the quadruped system.}
\vspace{-1.0em}
\label{fig:quad_diag} 
\end{center}
\end{figure}

%
%
We next apply our method for task-based, high-level mode planning on a real-world quadruped to reach a goal position in a terrain consisting of various obstacles (Fig. \ref{fig:exp_main_result}).
The quadruped has different walking modes, such as normal walking, slope walking, tilt walking, etc., each of which is modeled as a discrete mode in the hybrid system (see Fig. \ref{fig:quad_diag}). 
The obstacle set is assumed to be known \textit{a priori}; any continuous state going into this set in any discrete mode will immediately transition the system into frozen dynamics, so the reach BRT only captures the states from which the goal can be reached without any collisions.
Within each discrete mode, we use simplified continuous dynamics to describe center-of-mass evolution:
\begin{equation}
\begin{aligned}
\dot{x} & =k_x V_{N} \sin \varphi + d_x \\
\dot{y} & =k_y V_{N} \cos \varphi + d_y \\
\dot{\varphi} & =k_{\omega} \omega_{N},
\end{aligned}
\end{equation}
%
where the state $[x,y,\varphi]^T$ represent the global $xy$ position and heading angle of the robot's body. 
$V_{N} = 0.25m/s$ is the fixed (normalized) linear forward velocity and $\omega_{N} \in [-0.5, 0.5] rad/s$ is the angular velocity control of heading angle. 
$k_x$, $k_y$, and $k_{\omega}$ are velocity gains for different operation modes. 
$d_x$ and $d_y$ are bounded disturbances on $xy$ velocity. Also, we use discrete state $[z,\theta,\xi]^T$ to set the body's height, pitch, and roll angle in different operation modes for overcoming various obstacles. 

Each operation mode matches a specific obstacle or terrain shown in Fig. \ref{fig:exp_main_result}(b). The quadruped begins in mode $q_1$ for fast walking. If the robot reaches the boundary of the slope area, the system will be forced to switch to mode $q_2$ with a $20^{\circ}$ body pitch angle for slope climbing. While in normal walking mode, the control may switch to mode $q_3$ to walk slowly with a lower body height. This discrete mode allows the robot to make narrow turns or crawl under obstacles. If the robot goes into the boundary of a tilted obstacle, the system will go through a forced switch into mode $q_4$ to walk with a tilted body. Whenever the robot touches a ground obstacle, the system will stay in mode $q_5$ permanently with frozen dynamics.
The quadruped needs to reason about optimally switching between these different walking modes in order to reach its goal area (the $xy$ area marked as the pink cylinder in Fig. \ref{fig:exp_main_result}(a, b)) in the shortest time without colliding with any obstacles.

BRT for this experiment is calculated on a $[x,y,\varphi]$ grid of $[40,40,72]$ points over a time horizon of $T=45s$. \hl{The BRT calculation took 120 minutes on an AMD Ryzen7 4800H CPU.} Having the BRT, we can query the optimal velocity control, operation mode, and the desired discrete mode for all states within the time horizon. 
The quadruped is equipped with an Intel RealSense T265 tracking camera, allowing it to estimate its global state via visual-inertial odometry. 
Our framework's high-level optimal velocity control and discrete body states are then tracked by a low-level MPC controller that uses the centroidal dynamics of the quadruped \cite{kim2019highly}.

Our experiment results are shown in Fig. \ref{fig:exp_main_result} and can also be seen in the accompanying video.
In our setup, leveraging the slope on the right is the optimal (fastest) route to reach the target.
The corresponding trajectory is shown in orange in Fig. \ref{fig:exp_main_result}(a) and (b).
Specifically, the robot remains in the normal walking mode (blue boxes in Fig. \ref{fig:exp_main_result}(a)) and switches to the slope walking mode once near the slope (dark red boxes).
When approaching the end of the slope, it needs to make a tight turn to avoid a collision with the wall. 
Our framework can recognize that and transitions to the slow walking mode to allow for a tighter turn (green boxes). 
Once the turn is complete, the robot returns to normal walking mode. 
Fig. \ref{fig:exp_main_result}(c) shows first-person RGB images along the robot's path with color-coded number boxes to identify the active discrete mode.

We next put some papers on the slope for our second experiment in this scenario, making it more slippery. 
This change is encoded by adding a higher disturbance in the slope mode $q_2$. 
A new recalculated BRT marks the slope path as infeasible, and hence the robot needs to reach the target via the (slower) ground route.
The new robot trajectory is shown in 
light green in Fig. \ref{fig:exp_main_result}(a) and (b), and the first-person view in \ref{fig:exp_main_result}(d).
Once again, apart from selecting a new safe route, the proposed framework is able to switch between different walking modes along the route to handle tight turns and slanted obstacles, as shown by the activation of the tilt walking mode (cyan boxes in Fig. \ref{fig:exp_main_result}(a)).  
%
%

\begin{figure}[h] 
\begin{center} 
\includegraphics[trim=0 0 0 3em,clip,width=0.95\columnwidth]{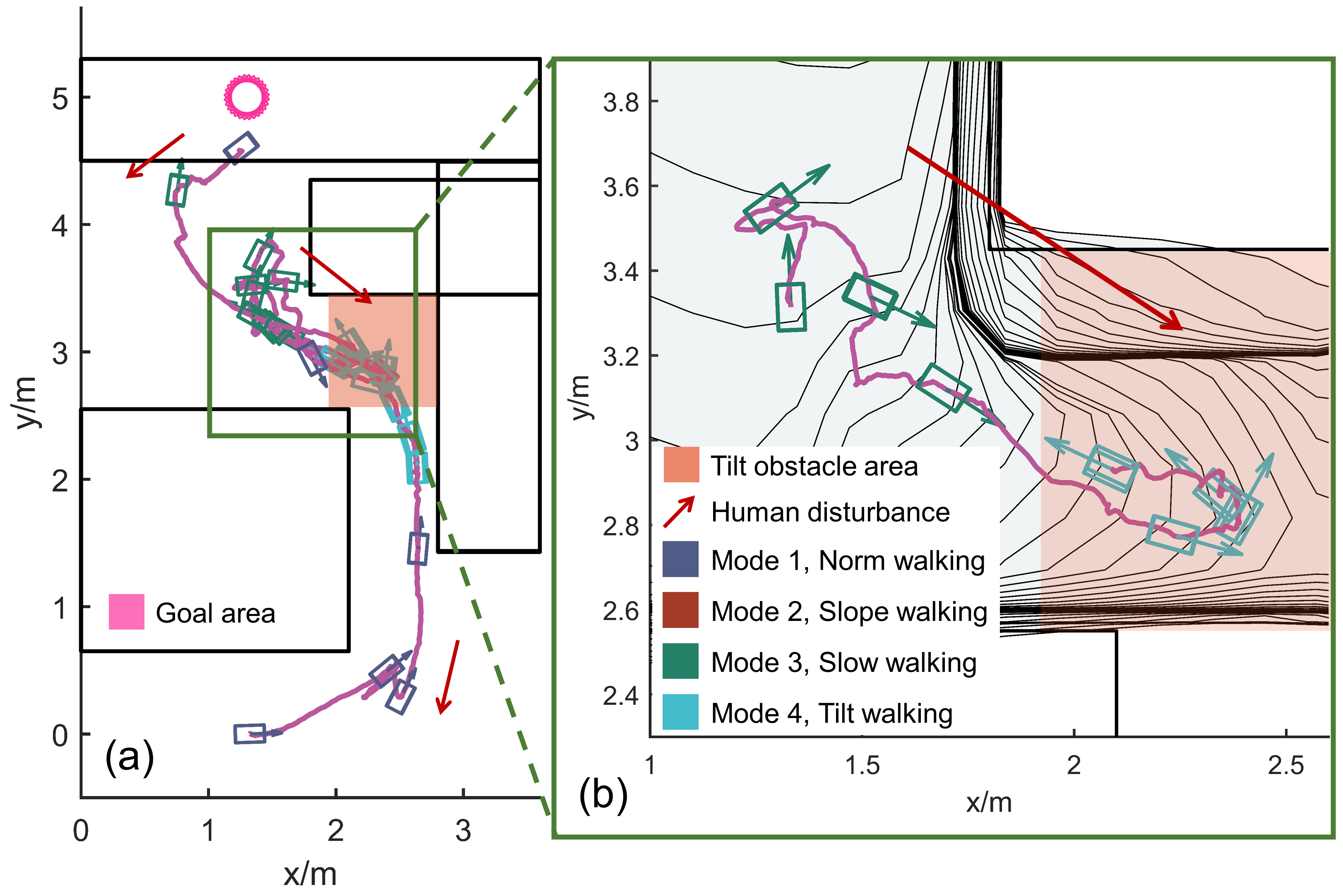}
\caption{(a) Robot trajectory with human disturbances. (b) Partial trajectory with overlaid reach BRT shows the robot autonomously switches from $q_3$ to $q_4$ 
after being pushed into the tilt obstacle area to ensure safety.}
\vspace{-1.5em}
\label{fig:traj_disturbance_1} 
\end{center}
\end{figure}

Finally, to test the closed-loop robustness of our method, we add human disturbance to the ground route experiment. 
The robot is kicked and dragged by a human during the locomotion process, as shown by red arrows in Fig. \ref{fig:traj_disturbance_1}. 
Specifically, the robot was dragged to face backward, pushing it closer to the tilted obstacle. 
However, the reachability controller is able to ensure safety by reactivating the tilt walking mode (Fig. \ref{fig:traj_disturbance_1}(b)), demonstrating that the proposed framework is able to reason about both closed-loop continuous \textit{and} discrete control inputs.

Nevertheless, it should be emphasized that the presented algorithm
only provides the top level of a hierarchical planning architecture that typically includes a robust footstep planner, a whole-body controller, and reliable state estimation \cite{norby2020fast}.
\hl{Given the extensive computational requirements of reachability-based methods, a particularly interesting future direction would be to use our framework as a high-level reference within high-fidelity global planners, such as RRT-Connect \cite{norby2020fast}, to reduce their sampling space and produce high-quality reference trajectories in real-time.}
\hl{Furthermore, the safety assurances are currently provided under perfect state estimation, an assumption that is often violated in the real world and results in accidental collisions during our experiments. Accounting for perception uncertainty within our framework is an important future research direction.}
%
%

%
\section{\label{conclusion}Discussion And Future Work}
\vspace{-0.5em}
We present an extension of the classical HJ reachability framework to hybrid dynamical systems with nonlinear dynamics, controlled and forced transitions, and state resets.
Along with the BRT, the proposed framework provides optimal continuous and discrete control inputs for ensuring the system safety and liveness.
Simulation studies and hardware experiments demonstrate the proposed method, both to reach goal areas and in maintaining safety.

Our work opens up several exciting future research directions. 
\hl{First, we rely on grid-based numerical methods to compute the BRT, whose computational complexity scales exponentially with the number of continuous states, limiting a direct use of our framework to relatively low-dimensional systems. 
We will explore recent advances in learning-based methods to solve high-dimensional HJI-VI \cite{bansal2021deepreach, AlbertDeepreachGuarantees, fisac2019bridging, ganai2024iterative, so2023solving} to overcome this challenge.} 
\hl{Another promising direction could be to explore alternative computational algorithms that leverage the structure in the target sets (e.g., convexity) and dynamics to efficiently compute the BRTs, such as Sum-of-Squares programming for polynomial dynamics \cite{landry2018reach} and Zonotope techniques for affine dynamics \cite{zono_poly_2010}.}
%
Second, in our current framework, we use freeze modes to 
avoid obstacle regions. 
In the future, we will extend our method to reach-avoid problems that naturally encode obstacle constraints within goal-reaching tasks.
Third, we currently assume perfect perception of the environment in our experiments. It would be interesting to explore how to use perception algorithms to automatically define different terrain regions and associated discrete modes in the hybrid system.
%
%
\hl{Fourth, it is important to ensure that the obtained safety controller is non-Zeno. Our current computation algorithm avoids Zeno behaviors by using a finite timestep $\delta$ for discrete transitions (automatically selected by the Level Set Toolbox to balance computational efficiency with approximation fidelity).
However, this comes at a price of approximate value functions and it would be interesting to explore how to properly handle Zeno behaviors during the reachable set and safety controller computations in Algorithm 1.} 
Finally, we are excited to apply our framework to a broader range of robotics applications and systems.
\section*{Acknowledgement}
\vspace{-0.5em}
This work is supported in part by the University of Santiago de Chile, Becas Chile, the NSF CAREER Program under award 2240163 and the DARPA ANSR program.


\bibliographystyle{plainnat}
\bibliography{./Bib/reachability, ./Bib/safe_motion_planning, ./Bib/bansal_papers, ./Bib/cbf_and_clf, ./Bib/mpc_based_safety, ./Bib/sial_bib}

\end{document}